\pdfoutput=1

\documentclass[10pt,twocolumn,letterpaper]{article}

\usepackage[pagenumbers]{cvpr}
\usepackage[pagebackref,breaklinks,colorlinks,citecolor=cvprblue]{hyperref}

\usepackage[T1]{fontenc}
\usepackage{microtype}
\usepackage{nicefrac}
\usepackage{multirow}
\usepackage{colortbl}
\usepackage{adjustbox}
\usepackage{pifont}
\usepackage{tikz}
\usetikzlibrary{arrows.meta, positioning, calc}
\providecolor{ForestGreen}{RGB}{34,139,34}
\providecolor{BrickRed}{RGB}{203,65,84}
\providecolor{camgreen}{RGB}{40,135,55}
\providecolor{camgrey}{RGB}{128,128,128}
\newcommand{\cgreen}{\textcolor{camgreen}{green}}
\newcommand{\cgrey}{\textcolor{camgrey}{grey}}
\providecolor{locours}{HTML}{2CA02C}
\providecolor{locpnp}{HTML}{C25E00}
\providecolor{locpnpf}{HTML}{D45BA5}
\providecolor{lochloc}{HTML}{0E8A99}
\providecolor{locgt}{RGB}{128,128,128}
\newcommand{\cmark}{\ding{51}}
\newcommand{\xmark}{\ding{55}}
\newcommand{\hc}[1]{\cellcolor{green!#1!red!25}#1}
\newcommand{\hcb}[1]{\cellcolor{green!#1!red!25}\textbf{#1}}
\newcommand{\hcu}[1]{\cellcolor{green!#1!red!25}\underline{#1}}
\newcommand{\gc}[1]{\cellcolor{gray!18}\textcolor{gray}{#1}}
\newcommand{\gcb}[1]{\cellcolor{gray!18}\textcolor{gray}{\textbf{#1}}}
\newcommand{\gcu}[1]{\cellcolor{gray!18}\textcolor{gray}{\underline{#1}}}
\newcommand{\better}[1]{\textcolor{ForestGreen}{#1}}
\newcommand{\worse}[1]{\textcolor{BrickRed}{#1}}
\newcommand{\ours}{DeepBA}

\providecolor{mapblue}{RGB}{91,132,196}
\newcommand{\cblue}{\textcolor{mapblue}{blue}}

\definecolor{cvprblue}{rgb}{0.21,0.49,0.74}

\newcommand{\huber}[2][\beta]{\operatorname{smooth}^{#1}_{L_1}\!\bigl(#2\bigr)}

\title{Deep Bundle Adjustment}

\author{{Pak Hong Chui}\\
Technical University of Munich\\
Munich Center for Machine Learning (MCML)\\
{\tt\small pak.chui@tum.de}
\and
Hector Andrade-Loarca\\
Technical University of Munich\\
Munich Center for Machine Learning (MCML)\\
{\tt\small Hector.Andrade@tum.de}
\and
Daniel Cremers\\
Technical University of Munich\\
Munich Center for Machine Learning (MCML)\\
{\tt\small cremers@tum.de}
}

\begin{document}
\twocolumn[{%
\renewcommand\twocolumn[1][]{#1}%
\maketitle
\begin{center}
    \captionsetup{type=figure}
    \includegraphics[width=\textwidth]{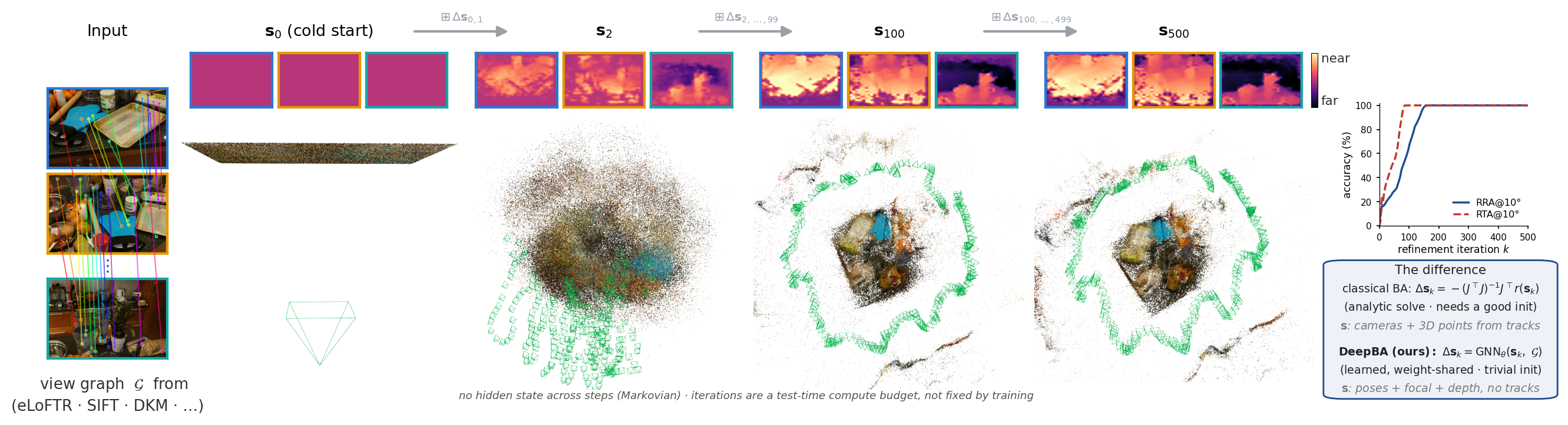}
    \captionof{figure}{\textbf{DeepBA runs the bundle-adjustment loop with a
    learned update, on a track-free state.} From a view graph $\mathcal{G}$ built by \emph{any}
    off-the-shelf matcher (\emph{Input}), a weight-shared GNN iteratively
    refines camera poses, intrinsics, and depth from a trivial cold start
    $\mathbf{s}_0$. The example is the full $240$-camera Mip-NeRF 360{~\cite{barron2022mipnerf360}}
    \texttt{counter} scene, with the estimated cameras in \cgreen{}.}
    \label{fig:teaser}
    \vspace{6pt}
\end{center}%
}]

\begin{abstract}
We present \textbf{DeepBA} (\textbf{Deep} \textbf{B}undle \textbf{A}djustment), a
graph neural network {(GNN)} that recovers camera poses, intrinsics, and scene structure from
2D matches across an arbitrary number of views. Rather than solving a
non-linear least-squares problem per scene, DeepBA amortizes bundle adjustment into
a learned graph network that iteratively refines its estimate from geometric
residuals. Where classical SfM computes its updates analytically and needs careful
initialization, DeepBA learns them and converges from a cold start. It is trained
with scale-invariant supervision (ground-truth rotation, translation direction,
and intrinsics, plus reprojection self-consistency), so datasets with arbitrary
metric units can be mixed without per-scene normalization. DeepBA is decoupled from the
matcher. At {2.04M} parameters, it outperforms prior learned SfM from
matches {on scene reconstruction and on most pose metrics},
is competitive with classical SfM at moderate accuracy thresholds and overtakes
it in wall time as the view count grows, and approaches monolithic image-to-geometry
transformers such as VGGT ($\sim$1.2B parameters) with greater modularity.
{Project~page: \mbox{\url{https://cphjsn.github.io/deep-ba/}}}
\end{abstract}

\section{Introduction}
Recovering camera poses and scene geometry from a set of images is a foundational problem in 3D computer vision, addressed by structure-from-motion (SfM)~\cite{schonberger2016sfm} and simultaneous localization and mapping (SLAM)~\cite{mur2015orbslam}. {Most modern SfM pipelines and keyframe-based SLAM systems} rely on the same classical solver, bundle adjustment (BA)~\cite{triggs1999bundle, hartley2003multiple}, which jointly refines cameras and 3D points by minimizing reprojection residuals via non-linear least squares. Bundle adjustment is near-optimal given a good initialization, but its non-convex landscape is fragile. {Incremental SfM obtains such an initialization through heuristics} and per-scene tuning~\cite{schonberger2016sfm}{, yet these heuristics break down} on weak baselines or planar structure~\cite{torr1999problem}. Meanwhile, learning-based {semi-dense~\cite{sun2021loftr} and dense~\cite{edstedt2023dkm, edstedt2024roma}} matchers have made matches far more reliable than SIFT-style keypoints{~\cite{lowe2004distinctive}}, shifting the bottleneck from matching to the solver. However, feeding them into a COLMAP-style pipeline does not automatically help, as its rigid {track building} is overwhelmed by the volume and its initialization stays brittle. This mismatch motivated a line of feed-forward 3D networks (DUSt3R~\cite{wang2024dust3r}, MASt3R~\cite{mast3r_eccv24}, and VGGT~\cite{vggt2024}) that regress geometry directly from images, trading the classical pipeline's modularity for robustness. There is a missing middle: a learned solver that \emph{amortizes} bundle adjustment itself, while leaving the matcher decoupled and solving any number of views at once. Classical bundle adjustment already operates on a graph of cameras and their 2D matches, so a graph neural network is the natural {architecture} for that structure. 

\vspace{+0.4cm}

We introduce \textbf{DeepBA} (\textbf{Deep} \textbf{B}undle \textbf{A}djustment, \cref{fig:teaser}), a GNN that takes 2D matches from an off-the-shelf matcher and jointly infers absolute camera poses, intrinsics, and 3D points. DeepBA replaces the per-scene non-linear least-squares solver with a learned {Markovian} iterative refinement. {Each iteration recomputes} geometric features from the current state and {predicts} an updated state via shared GNN weights. The number of iterations is a free choice at test time, independent of the training schedule: {accuracy rises with the iteration count until it saturates}, mirroring an iterative classical solver. It operates directly on pairwise matches, with no separate track-building stage. Our contributions are:
\begin{itemize}
    \item \textbf{Amortized {cold-start} bundle adjustment via a GNN.} We formulate bundle adjustment as a learned iterative refinement over the view graph of any number of views, with a Markovian update that lets the user trade {runtime} for accuracy at test time. {It requires no initial poses or intrinsics and converges from a cold start, where classical BA fails.}
    \item \textbf{Scale-invariant, anchor-free, multi-dataset supervision.} Chordal rotation, direction-only translation, intrinsics, self-consistent reprojection and 3D losses train DeepBA on datasets with arbitrary
    metric units without per-scene normalization. Because DeepBA predicts state updates rather than geometry, it needs no anchor camera, unlike VGGT~\cite{vggt2024}, which uses the first camera as its reference frame.
    \item \textbf{Robustness across matcher families.} \ours{} sees only 2D matches, so a trained network can run from sparse to dense matching without retraining. We demonstrate this extensively, across every matcher family.
    \item \textbf{Competitive accuracy at a fraction of the size.} With {$2.04$M} parameters ({${\sim}71\times$} fewer than the strongest prior match-based network, GASfM{~\cite{brynte2024learning}} at $145$M), DeepBA outperforms learned SfM from matches {on most metrics} and is competitive with classical SfM at moderate thresholds, approaching image-to-geometry transformers such as VGGT (${\sim}1.2$B parameters).
\end{itemize}
\section{Related Work}

\paragraph{{Classical structure-from-motion and bundle adjustment.}}
The classical pipeline for multi-view reconstruction proceeds by feature
extraction, pairwise matching, geometric verification, incremental pose estimation, and a final non-linear
refinement via {BA}~\cite{triggs1999bundle, hartley2003multiple}.
COLMAP~\cite{schonberger2016sfm} is its standard implementation and the baseline
most work compares against. GLOMAP~\cite{pan2024glomap} replaces its incremental stage
with global rotation averaging followed by a positioning step that solves for all
cameras and points at once. That drops the seed-and-order heuristics, but leans
on the pairwise rotations instead.
Bundle adjustment's accuracy hinges on initialization~\cite{lourakis2009sba}
and hand-tuned heuristics, making it brittle on weak baselines, planar or
low-texture scenes, and repetitive structure~\cite{torr1999problem, he2024dsfm, heinly2014correcting}. A long line of work has widened bundle adjustment's basin of attraction with alternative
error parameterizations and convex relaxations~\cite{Hong2016projba, Zach2018pose, iglesias2023expose, weber2024power, han2025building}, which improves
cold-start convergence. These remain per-scene optimizations and still {face}
the mirror ambiguity and the global-scale gauge freedom of unconstrained
multi-view geometry~\cite{strasdat2010scale, nister2004efficient}. Recent neural pipelines inherit the same structure, still running an optimization once per scene at test time. {For example,} Detector-Free SfM~\cite{he2024dsfm} adapts track building to dense matchers, and FlowMap~\cite{smith2024flowmap} fits poses, intrinsics, and depth by gradient descent through a dense-flow objective. {TFD-SfM~\cite{chui2027tfdsfm} removes track building altogether. It lifts both pixels of every correspondence to their own 3D landmarks and optimizes the reconstruction from a cold start.} {Similar to TFD-SfM,} DeepBA {is track-free} but amortizes the per-scene {optimization} into {a} learned refinement from {a} cold start.

\paragraph{Feed-forward 3D reconstruction.}
Recent works remove the classical pipeline, training neural
networks to regress geometry directly from images.
PoseDiffusion~\cite{wang2023posediffusion} diffuses camera poses alone. DUSt3R~\cite{wang2024dust3r}
and MASt3R~\cite{mast3r_eccv24} produce per-camera pointmaps in a shared frame, with
MASt3R adding an explicit matching head for downstream localization. Both are
pair-only: extending them to many views needs a
global-alignment step (e.g.\ MASt3R-SfM~\cite{duisterhof2025mast3rsfm}) that
re-introduces a per-scene non-linear {optimization}. VGGSfM~\cite{wang2024vggsfm} combines
deep pose and triangulator modules with a differentiable bundle adjustment back-end, while
VGGT~\cite{vggt2024} pushes furthest, absorbing the matcher, visual
encoder, and geometry solver into a single transformer that ingests
dozens of views in one forward pass, at the cost of model
size, training-data scale, and tight perception--geometry coupling. \ours{} instead operates on
matches rather than image features and reasons jointly over all views,
needing no fusion of pairwise reconstructions. Keeping the matcher
external rather than learning one, it is {${\sim}590{\times}$} smaller than VGGT and
trainable on substantially less data.

\paragraph{Iterative learned refinement.}
\looseness=-1
RAFT~\cite{teed2020raft} introduced iterative refinement with a fixed update
network applied for a variable number of {iterations}, recomputing features from the
current estimate at each {one}. The template has been extended to stereo,
scene flow, and visual SLAM~\cite{lipson2021raft,teed2021raft,teed2021droid}.
What makes the template work is supervising the intermediate iterations. The network
cannot then rely on one large
correction, and instead learns a contraction, an update that improves the
estimate every time it is applied. The number of iterations therefore becomes a
test-time trade between accuracy and {runtime}.
DeepBA carries this template into the multi-view geometry setting with a
weight-shared GNN (a Markovian update, without a recurrent gate). The closest
learned refiner is DROID-SLAM~\cite{teed2021droid}, which targets calibrated,
temporally-ordered video with a coupled dense feature network, whereas DeepBA handles
unordered, wide-baseline collections and recovers intrinsics from a trivial
initialization.

\paragraph{Learning-based SfM from matches.}
\looseness=-1
The work closest to ours also learns SfM directly from 2D matches
rather than from pixels. Most of it regresses the reconstruction with a network and ends in a classical
bundle adjustment. Deep Permutation Equivariant SfM~\cite{moran2021deep} and its successor RESfM~\cite{khatib2025resfm} regress cameras and sparse structure
by minimizing reprojection error. Both fine-tune the network on each test scene before the final bundle adjustment.
GASfM~\cite{brynte2024learning} instead replaces the equivariant layers of~\cite{moran2021deep} with graph attention over the same point tracks. It is feed-forward, restricted to a calibrated or purely
projective regime, and refined by a final classical bundle adjustment. {LS-Net~\cite{clark2018lsnet} learns a non-linear least-squares solver for dense, calibrated two-view stereo.
Learning-to-Bundle-Adjust~\cite{tanaka2021learning} predicts BA updates with a graph network, refining the local map of an initialized vehicular SLAM system.
BA-Net~\cite{tang2018ba} starts its poses at the identity but runs a dense feature-metric BA on calibrated short clips, with a learned {Levenberg--Marquardt} damping and an image-dependent depth basis.} PoGO-Net~\cite{li2021pogo} is a graph network for rotation/motion averaging that consumes
relative poses, not raw matches, and recovers only orientations. {Unlike these methods, \ours{} learns the update of poses, intrinsics and depth itself. It recomputes geometric features of the matches from its current estimate at every refinement iteration. It starts every scene from the same initial state.}
\section{Method}

\begin{figure*}[t]
    \centering
    \includegraphics[width=0.92\linewidth]{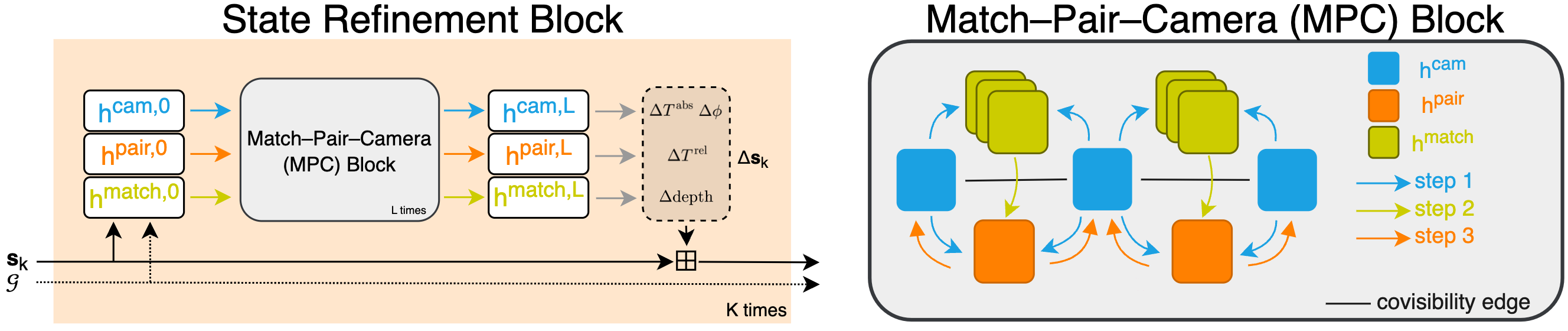}
    \caption{\textbf{DeepBA architecture.}
    \emph{Left:}~one refinement iteration: the scene state {$\mathbf{s}_k$} and
    view graph $\mathcal{G}$ are encoded into {$(\mathbf{h}^{\text{pair}}, \mathbf{h}^{\text{match}})$}, re-initialized every iteration
    (Markovian), mixed by $L$ MPC blocks, and decoded into the state delta
    $\Delta\mathbf{s}_k$ composed via $\boxplus$ (\cref{eq:boxplus}); the
    block is weight-shared across the $K$ iterations.
    \emph{Right:}~one MPC block; its {three} message-passing steps are
    specified in \cref{sec:method:network}.}
    \label{fig:arch}

\end{figure*}

We introduce \textbf{DeepBA}, a graph neural network that performs amortized bundle adjustment via iterative refinement over a graph of cameras and their pairwise matches (\cref{fig:arch}).

\subsection{Problem Setting and Notation}
\label{sec:method:problem}

\looseness=-1
Given $N$ unordered images of a static scene, DeepBA maps pairwise 2D matches to per-image estimates of absolute pose, intrinsics, and a depth map (3D points are recovered by unprojection). The matches are image-point pairs from an off-the-shelf matcher. The inputs form a \emph{view graph} $\mathcal{G} = (\mathcal{V}, \mathcal{E}_{\text{pair}}, \mathcal{E}_{\text{match}})$ in which $\mathcal{V}$ is the set of $N$ cameras, $\mathcal{E}_{\text{pair}}$ is the set of $P$ pairs of cameras connected by at least $50$ matches, and $\mathcal{E}_{\text{match}}$ is the set of {all} $M$ matches {over all pairs of the view graph}.
Each camera $i \in \mathcal{V}$ carries an absolute pose $T^{\text{abs}}_i$, a {log field of view (log-FOV)} $\phi_i$ (our default camera model, more detail in \cref{app:intrinsics}), and a log-depth map $D_i \in \mathbb{R}^{1 \times H_d \times W_d}$ on a fixed grid with $H_d = W_d = 32$. Each pair $(i, j) \in \mathcal{E}_{\text{pair}}$ carries a relative pose $T^{\text{rel}}_{ij} = (\hat{\mathbf{t}}^{\text{rel}}_{ij}, \boldsymbol{\omega}^{\text{rel}}_{ij})$ in a \emph{decoupled} parameterization: a unit translation direction $\hat{\mathbf{t}}^{\text{rel}}_{ij} \in S^2$ and a rotation vector $\boldsymbol{\omega}^{\text{rel}}_{ij} \in \mathbb{R}^3$. Together, these form the \emph{scene state}
\begin{equation}
    \mathbf{s} \;:=\; \bigl(\{T^{\text{abs}}_i\}_{i \in \mathcal{V}},\;
    \{T^{\text{rel}}_{ij}\}_{(i,j) \in \mathcal{E}_{\text{pair}}},\;
    \{\phi_i\}_{i \in \mathcal{V}},\;
    \{D_i\}_{i \in \mathcal{V}}\bigr).
    \label{eq:scene-state}
\end{equation}
{The absolute poses, FOVs and depth maps already define the reconstruction. We still keep a relative pose per pair, which the network refines at every iteration under {ground-truth (GT)} supervision, while the absolute poses receive no GT supervision. The motivation of this design comes from pose-graph optimization. The absolute poses are updated to agree with the current relative poses instead of being pulled directly toward the GT, so they follow the gradual refinement of the relative poses. A relative pose is only a rotation and a translation direction, so its GT supervision needs no metric scale, while the reprojection error sets the translation lengths of the absolute poses.}
At every refinement iteration, the network predicts a delta
\begin{equation}
\begin{split}
    \Delta\mathbf{s} \;=\; \bigl(&\{\Delta T^{\text{abs}}_i\}_{i \in \mathcal{V}},\;
    \{\Delta T^{\text{rel}}_{ij}\}_{(i,j) \in \mathcal{E}_{\text{pair}}},\\
    &\{\Delta \phi_i\}_{i \in \mathcal{V}},\;
    \{\Delta D_i\}_{i \in \mathcal{V}}\bigr),
\end{split}
\label{eq:state-delta}
\end{equation}
with $\Delta T^{\text{abs}}_i \in \mathbb{R}^6$ in Lie-algebra coordinates, and
composes it onto the current state,
\begin{equation}
    \mathbf{s}_{k+1} \;=\; \mathbf{s}_k \,\boxplus\, \Delta\mathbf{s}_k,
    \label{eq:boxplus}
\end{equation}
where $\boxplus$ acts component-wise (exact update rules and conventions in \cref{app:conventions}).

\subsection{Per-Iteration Geometric Features}
\label{sec:method:features}

\looseness=-1 At every refinement iteration, DeepBA computes three hidden states from scratch (\emph{Markovian}: no hidden state crosses iterations): the per-pair $\mathbf{h}^{\text{pair}}$ and per-match $\mathbf{h}^{\text{match}}$ encode closed-form \emph{geometric features} of the current scene state, while the per-camera $\mathbf{h}^{\text{cam}}$ takes no geometric input and starts from a single learned embedding $\mathbf{e}^{\text{cam}} \in \mathbb{R}^{d_h}$ shared by every camera. {All three hidden states have dimension $d_h=128$.}
\subsubsection{{Pair Geometric Features} ($\rightarrow \mathbf{h}^{\text{pair}}$)}
\label{sec:method:features:pair}

{The pair geometric features give the network pairwise information.} Although a pair stores a single relative pose, we featurize it in both directions {so that message passing in the GNN reaches both of its cameras}: we construct a $14$-dim feature per direction, $\mathbf{f}^{\text{pair}}_{ij}$ for the forward edge ($i \rightarrow j$) and $\mathbf{f}^{\text{pair}}_{ji}$ for the inverse ($j \rightarrow i$), and encode each into its own per-pair hidden state, $\mathbf{h}^{\text{pair}}_{ij}$ and $\mathbf{h}^{\text{pair}}_{ji}$. The forward feature is
\begin{equation}
    \mathbf{f}^{\text{pair}}_{ij} \;=\; \bigl[\,
    \underbrace{\hat{\mathbf{t}}^{\text{rel}}_{ij},\, \boldsymbol{\omega}^{\text{rel}}_{ij}}_{\text{rel-pose state}\;(6)} \,,\;
    \underbrace{\phi_i,\, \phi_j}_{\text{intrinsics}\;(2)} \,,\;
    \underbrace{\boldsymbol{\delta}^{\text{t}}_{ij},\; \boldsymbol{\delta}^{\text{R}}_{ij}}_{\text{{pose-graph} residual}\;(6)}
    \,\bigr] %
\end{equation}
where the \emph{pose-graph residual} measures the disagreement between the directly predicted relative pose $T^{\text{rel}}_{ij}$ and the counterpart $T^{\text{abs}}_{ij}$ derived from the two absolute poses (exact form in \cref{app:pose-graph}).

\subsubsection{{Match Geometric Features} ($\rightarrow \mathbf{h}^{\text{match}}$)}
\label{sec:method:features:match}

{The match geometric features tell the network how well the current state explains each match.} Each match $e_{ij} \in \mathcal{E}_{\text{match}}$, with endpoints $\mathbf{u}^{e_{ij}}_{\text{src}}, \mathbf{u}^{e_{ij}}_{\text{dst}} \in \mathbb{R}^2$ in images $i$ and $j$, contributes a $12$-dim feature per direction---$\mathbf{f}^{\text{match}}_{e_{ij}}$ for the forward edge ($i \rightarrow j$) and $\mathbf{f}^{\text{match}}_{e_{ji}}$ for the inverse---and is encoded into its own per-match hidden state $\mathbf{h}^{\text{match}}_{e_{ij}}$ and $\mathbf{h}^{\text{match}}_{e_{ji}}$. For brevity, we drop the edge subscript and write $e := e_{ij}$ throughout the remainder of this subsection. The forward feature is
\begin{equation}
    \mathbf{f}^{\text{match}}_{e} \;=\; \bigl[\,
    \underbrace{\mathbf{u}^{e}_{\text{src}},\, \mathbf{u}^{e}_{\text{dst}}}_{4} \,,\;
    \underbrace{\boldsymbol{\epsilon}_{e}}_{2} \,,\;
    \underbrace{\boldsymbol{\rho}_{e}}_{2} \,,\;
    \underbrace{\log\|\mathbf{P}^{e}_{\text{dst}}\|}_{1} \,,\;
    \underbrace{\boldsymbol{\delta}^{\text{3D}}_{e}}_{3}
    \,\bigr] %
    \label{eq:match-feat}
\end{equation}
Beyond $(\mathbf{u}^{e}_{\text{src}},\, \mathbf{u}^{e}_{\text{dst}})$, the remaining entries are closed-form functions of the current state {(full formulas in \cref{app:match-feats}). The} \emph{epipolar vector} $\boldsymbol{\epsilon}_{e}$ {is} the point-to-epipolar-line offset under the {directly predicted} relative pose{. The others are} the \emph{2D reprojection residual} $\boldsymbol{\rho}_{e}${, the} \emph{3D consistency residual} $\boldsymbol{\delta}^{\text{3D}}_{e}$, and the log-magnitude $\log\|\mathbf{P}^{e}_{\text{dst}}\|$ of the destination point.

\subsection{Iterative Refinement Network}
\label{sec:method:network}

\paragraph{State refinement block.}
At iteration $k = 0, \dots, K-1$, the \emph{state refinement block} (\cref{fig:arch}, left) runs three steps.

\begin{itemize}[leftmargin=1.2em,itemsep=2pt,topsep=2pt]
    \item \textbf{Step 1 (encode).} The features $\mathbf{f}^{\text{pair},k}, \mathbf{f}^{\text{match},k}$ enter per-type hidden states through a linear map,
    \begin{equation}
        \mathbf{h}^{\bullet,0} \;=\; W^{\bullet}\,\mathbf{f}^{\bullet,k}, \quad \bullet \in \{\text{pair}, \text{match}\},
        \label{eq:encode}
    \end{equation}
    while $\mathbf{h}^{\text{cam},0}$ is set to the learned embedding $\mathbf{e}^{\text{cam}}$.
    \item \textbf{Step 2 (mix).} A stack of $L = 4$ Match--Pair--Camera (MPC) blocks mixes the three tiers over the graph,
    \begin{equation}
        \bigl(\mathbf{h}^{\text{cam}}, \mathbf{h}^{\text{pair}}, \mathbf{h}^{\text{match}}\bigr) \;\leftarrow\; \mathrm{MPC}_\ell\bigl(\mathbf{h}^{\text{cam}}, \mathbf{h}^{\text{pair}}, \mathbf{h}^{\text{match}};\, \mathcal{G}\bigr),
        \label{eq:mpc-stack}
    \end{equation}
    for $\ell = 1, \dots, L$. The $L$ blocks do not share weights.
    \item \textbf{Step 3 (decode).} The prediction heads emit the state delta $\Delta\mathbf{s}_k$.
\end{itemize}

\paragraph{MPC block composition.}
Each MPC block (\cref{fig:arch}, right) updates the three hidden states in {three} steps. 

\begin{itemize}[leftmargin=1.2em,itemsep=2pt,topsep=2pt]
    \item \textbf{Step 1 (camera$\to$pair/match broadcast).} Injects endpoint camera states into per-pair and per-match channels:
    \begin{equation}
    \begin{aligned}
        \mathbf{h}^{\text{pair}}_{ij} &\;\mathrel{+}=\; W^{\text{pair}}_{\text{1}}\!\bigl(\mathbf{h}^{\text{cam}}_{i}\,\Vert\,\mathbf{h}^{\text{cam}}_{j}\bigr), \\
        \mathbf{h}^{\text{match}}_{e_{ij}} &\;\mathrel{+}=\; W^{\text{match}}_{\text{1}}\!\bigl(\mathbf{h}^{\text{cam}}_{i}\,\Vert\,\mathbf{h}^{\text{cam}}_{j}\bigr),
    \end{aligned}
    \end{equation}
    where $W^{\bullet}$ represents the learned linear map $\mathbb{R}^{2d_h}\!\to\!\mathbb{R}^{d_h}$.
    \item \textbf{Step 2 (match$\to$pair aggregation).} Matches are conditioned on their parent pair and aggregated into it by attention over the pair's matches $\mathcal{E}_{ij}$:
    \begin{equation}
    \begin{aligned}
        \mathbf{h}^{\text{match}}_{e} &\;\mathrel{+}=\; W^{\text{match}}_{\text{2}}\bigl(\mathbf{h}^{\text{match}}_{e} \,\Vert\, \mathbf{h}^{\text{pair}}_{ij}\bigr), \\
        \mathbf{h}^{\text{pair}}_{ij} &\;\mathrel{+}=\; \mathrm{Attn}\bigl(\mathbf{h}^{\text{pair}}_{ij},\, \{\mathbf{h}^{\text{match}}_{e}\}_{e \in \mathcal{E}_{ij}}\bigr),
    \end{aligned}
    \end{equation}
    where $\mathrm{Attn}(\mathbf{q}, \mathcal{S})$ is $4$-head dot-product attention~\cite{vaswani2017attention} (query $\mathbf{q}$, keys/values from $\mathcal{S}$).
    \item \textbf{{Step 3} (pair$\to$camera attention).} Each camera attends over its incident directional pair states{.\ The} forward $(i{\to}j)$ state reaches camera $j$, the inverse $(j{\to}i)$ state camera $i$:
    \begin{equation}
        \mathbf{h}^{\text{cam}}_{j} \;\mathrel{+}=\; \mathrm{Attn}\bigl(\mathbf{h}^{\text{cam}}_{j},\, \{\mathbf{h}^{\text{pair}}_{ij}\}_{i \in \mathcal{N}(j)}\bigr),
    \end{equation}
    with $\mathcal{N}(j)$ the cameras paired with $j$.
\end{itemize}

\paragraph{Prediction heads.}
Four heads read the hidden states of the final MPC block and emit the change of the state $\Delta\mathbf{s}_{k}$.

\begin{itemize}[leftmargin=1.2em,itemsep=2pt,topsep=2pt]
    \item \textbf{Absolute-pose head} and \textbf{log-FOV head.} Two MLPs on $\mathbf{h}^{\text{cam}}$ produce $\Delta T^{\text{abs}}_i \in \mathbb{R}^6$ and $\Delta\phi_i \in \mathbb{R}$, respectively.
    \item \textbf{Relative-pose head.} An MLP on $\mathbf{h}^{\text{pair}}_{ij}$ produces $\Delta T_{ij} \in \mathbb{R}^6$ in the decoupled $[\hat{\mathbf{t}},\boldsymbol{\omega}]$ parameterization. The change of relative pose is obtained by
    \begin{equation}
        \Delta T^{\text{rel}}_{ij}
        \;=\; \tfrac{1}{2}\bigl(\Delta T_{ij}
            - \Delta T_{ji}\bigr).
        \label{eq:antisym-pose}
    \end{equation}
    {Here $\Delta T_{ji}$ is the output of the same head on the reversed state $\mathbf{h}^{\text{pair}}_{ji}$ and predicts the change of $T^{\text{rel}}_{ji} = (T^{\text{rel}}_{ij})^{-1}$. {We use $-\Delta T_{ji}$ as a second prediction of the change of $T^{\text{rel}}_{ij}$.} \cref{eq:antisym-pose} averages the two predictions, which makes the update antisymmetric by construction, $\Delta T^{\text{rel}}_{ji} = -\Delta T^{\text{rel}}_{ij}$.}
    \item \textbf{Depth head.} A small spatial decoder: each $\mathbf{h}^{\text{match}}$ is scattered onto its destination camera's $H_d{\times}W_d$ grid at the match location, pooled by a {bilinear} per-cell mean, and decoded into the log-depth delta $\Delta D_i$ (details in \cref{app:depth-head}).
\end{itemize}

\subsection{Training Loss}
\label{sec:method:training}

\paragraph{Scale-invariant pose supervision.}
We define a scale-invariant pose distance that combines a chordal rotation distance with a translation-direction cosine,
\begin{equation}
    d\bigl(T_a, T_b\bigr) \;=\; \underbrace{\bigl(1 - \cos\theta_R\bigr)}_{\text{rotation cosine on }SO(3)} \;+\; \underbrace{\bigl(1 - \cos\theta_{\mathbf{t}}\bigr)}_{\text{translation cosine on }S^2},
    \label{eq:pose-dist}
\end{equation}
where $\theta_R \in [0, \pi]$ is the geodesic angle between $R_a$ and $R_b$ on $SO(3)$, computed from the relative rotation $R_a R_b^\top$ via $\cos\theta_R = \tfrac{1}{2}\bigl(\mathrm{tr}(R_a R_b^\top) - 1\bigr)$, and $\theta_{\mathbf{t}} \in [0, \pi]$ is the angle between the unit-direction translations on $S^2$, $\cos\theta_{\mathbf{t}} = \langle \hat{\mathbf{t}}_a, \hat{\mathbf{t}}_b \rangle$ with $\hat{\mathbf{t}} := \mathbf{t}/\|\mathbf{t}\|$. Because $d\bigl(T_a, T_b\bigr)$ never consumes the magnitude of $\mathbf{t}$, no metric scale leaks into supervision. This {enables} us to mix datasets with arbitrary metric units without per-scene normalization. For every pair $(i, j)$, we define two different residuals,
\begin{equation}
    r^{\text{rel}}_{ij}  \;=\; d\!\bigl(T^{\text{rel}}_{ij},\, T^{\text{gt}}_{ij}\bigr), \qquad
    r^{\text{abs}}_{ij} \;=\; d\!\bigl(T^{\text{abs}}_{ij},\, \mathrm{sg}\!\bigl(T^{\text{rel}}_{ij}\bigr)\bigr),
    \label{eq:r-two}
\end{equation}
where $\mathrm{sg}(\cdot)$ is the stop-gradient operator. $r^{\text{rel}}$ supervises the relative-pose head against GT{. The term} $r^{\text{abs}}$ couples the absolute-pose head to the (GT-supervised) relative-pose head, supervising the absolute poses through multi-view consensus, a learned analogue of pose-graph optimization. {The stop-gradient makes the absolute poses follow the relative poses, never the reverse.} The pose loss averages both over all pairs:
\begin{equation}
    \mathcal{L}_{\text{pose}} \;=\; \frac{1}{|\mathcal{E}_{\text{pair}}|} \sum_{(i,j) \in \mathcal{E}_{\text{pair}}} \bigl(r^{\text{rel}}_{ij} + r^{\text{abs}}_{ij}\bigr).
    \label{eq:pose}
\end{equation}

\paragraph{Geometric losses.}
We further enforce self-consistency between the model's outputs through a geometric reprojection loss that requires no ground-truth scale. Writing $\huber{\cdot}$ for the smooth-$L_1$ kernel~\cite{huber1964robust,girshick2015fast} of width $\beta$ (\cref{app:impl}), the 2D reprojection and 3D consistency losses average the robustified residual norms over all matches $e \in \mathcal{E}_{\text{match}}$:
\begin{equation}
    \begin{aligned}
        \mathcal{L}_{\text{2d}} &= \frac{1}{|\mathcal{E}_{\text{match}}|} \sum_{e} w_e \huber[0.01]{\|\boldsymbol{\rho}_{e}\|}, \\
        \mathcal{L}_{\text{3d}} &= \frac{1}{|\mathcal{E}_{\text{match}}|} \sum_{e} w_e \huber[0.01]{\|\boldsymbol{\delta}^{\text{3D}}_{e}\|},
    \end{aligned}
\end{equation}
where $\boldsymbol{\rho}_{e}$ and $\boldsymbol{\delta}^{\text{3D}}_{e}$ are the per-match residuals of \cref{sec:method:features:match} and $w_e = \gamma^2 / (\gamma^2 + s^2_e) \in (0, 1]$ is a Cauchy weight on match $e$, held constant during backpropagation, with $s_e$ the match's Sampson distance under the GT epipolar geometry (kernel width $\gamma = 5 \times 10^{-3}$). As a result, $\mathcal{L}_{\text{2d}}/\mathcal{L}_{\text{3d}}$ are effectively evaluated on inliers only, encouraging the network to fit the correct matches and to learn to ignore the inaccurate ones. 

The FOV loss applies the same kernel to the log-FOV error against GT,
\begin{equation}
    \mathcal{L}_{\text{fov}} \;=\; \frac{1}{|\mathcal{V}|} \sum_{i \in \mathcal{V}} \huber[1]{\phi_i - \phi_i^{\text{gt}}}.
\end{equation}

\paragraph{Final loss.}
We run a Markov trajectory of $K_{\text{train}}$ iterations per training step and {backpropagate} through 24 of them ($\mathcal{K}_{{\text{grad}}}$, \cref{app:impl}). In practice, even after the reconstruction has converged, the refiner keeps emitting updates that are pure global rotations and translations of the scene. We therefore add a \emph{fixed-point} regularizer $\mathcal{L}^k_{\text{fp}} = \|\Delta\mathbf{s}_k\|_1$ that penalizes the magnitude of these unnecessary, gauge-equivalent updates, pressing late iterations toward true stationarity.
The total loss sums over a view graph,
\begin{equation}
    \begin{aligned}
    \mathcal{L}_{\text{final}} = \!\!\sum_{k \in \mathcal{K}_{\text{grad}}}\!\! \Bigl(
    &\underbrace{\mathcal{L}^k_{\text{pose}} + \mathcal{L}^k_{\text{fov}}}_{\text{supervised}}
    \;+\;
    \underbrace{\lambda_{\text{2d}} \mathcal{L}^k_{\text{2d}} + \lambda_{\text{3d}} \mathcal{L}^k_{\text{3d}}}_{\text{{self-consistency}}} \\[1pt]
    &\;+\;
    \underbrace{\lambda_{\text{fp}} \mathcal{L}^k_{\text{fp}}}_{\text{regularizer}}
    \Bigr).
    \end{aligned}
    \label{eq:final-loss}
\end{equation}
\section{Experiments}
\label{sec:experiments}

We evaluate DeepBA on multi-view camera pose across six benchmarks (\cref{sec:exp:pose}) and on structure accuracy on DTU (\cref{sec:exp:structure}), then ablate and analyze the upstream matcher, network capacity, outlier robustness, generalization beyond the training budget, and runtime scaling (\cref{sec:exp:ablation}). The appendix adds the accuracy-versus-iteration profile (\cref{app:convergence}) and testing of camera localization on a frozen \ours{} map (\cref{app:localization}). In this paper, we treat all image collections as unordered and without known intrinsics unless explicitly noted.

\subsection{Setup}
\label{sec:exp:setup}

\paragraph{Evaluation datasets.}
Following prior work~\cite{wang2024dust3r,mast3r_eccv24,vggt2024,wang2024vggsfm}, we evaluate on six benchmarks that span diverse imaging conditions: CO3Dv2~\cite{reizenstein21co3d}, RealEstate10K~\cite{zhou2018stereo}, ETH3D~\cite{schops2017multi}, DTU~\cite{Jensen_2014_CVPR}, Mip-NeRF 360~\cite{barron2022mipnerf360}, and IMC2021~\cite{jin2021image}. Per-dataset scene selection and frame sampling are detailed in \cref{app:datasets}. Five benchmarks (RealEstate10K, ETH3D, DTU, Mip-NeRF 360, and IMC2021) are absent from our training mixture, so they measure \emph{cross-dataset} generalization; CO3Dv2 is in the mixture but is evaluated only on held-out scenes disjoint from the training set.

\paragraph{Training data.}
DeepBA is trained on a heterogeneous mixture of 3D datasets: MegaDepth~\cite{megadepth18}, CO3Dv2~\cite{reizenstein21co3d}, ScanNet~\cite{dai2017scannet}, BlendedMVS~\cite{yao2020blendedmvs}, DL3DV~\cite{ling2024dl3dv}, and Calle~\cite{olsson2011stable,enqvist2011nonsequential}, spanning indoor/outdoor, object- and scene-centric captures. Their arbitrary and incompatible metric units mix safely under DeepBA's scale-invariant supervision (\cref{sec:method:training}), without per-scene normalization.

\paragraph{Baselines.}
We compare against three families: \emph{classical SfM} (COLMAP~\cite{schonberger2016sfm}, GLOMAP~\cite{pan2024glomap}); \emph{learned image-to-geometry} (PoseDiffusion~\cite{wang2023posediffusion}, DUSt3R~\cite{wang2024dust3r}, MASt3R~\cite{mast3r_eccv24}, MASt3R-SfM~\cite{duisterhof2025mast3rsfm}, VGGSfM~\cite{wang2024vggsfm}, VGGT~\cite{vggt2024}); and \emph{learning-based SfM from matches} (GASfM~\cite{brynte2024learning}, RESfM~\cite{khatib2025resfm}).

\paragraph{Metrics.}
For pose estimation, following~\cite{vggt2024,mast3r_eccv24}, we report Relative Rotation/Translation Accuracy (RRA/RTA) and AUC@30\textdegree{}, the area under the joint accuracy curve up to 30\textdegree{}. For structure quality, we report Accuracy and Completeness against DTU's structured-light reference scans at $1/2/5$\,cm. Both follow the standard definitions, given in \cref{app:metrics}.

\paragraph{Implementation.}
DeepBA uses $L{=}4$ MPC blocks ({$2.04$M} parameters), trained for $50{,}000$ steps with AdamW{~\cite{loshchilov2019decoupled}} (peak learning rate {$6{\times}10^{-4}$}) and loss weights {$\lambda_{\text{2d}}{=}1$}, $\lambda_{\text{3d}}{=}\lambda_{\text{fp}}{=}0.1$, on {4 NVIDIA L40S} GPUs. Each step unrolls the refiner for all $K_{\text{train}}{=}100$ iterations but {backpropagates} only through a $24$-iteration subset. The scene state is initialized input-independently (identity poses, zero log-FOV and log-depth) and perturbed with Gaussian noise during training. 
{Unless stated otherwise} (\cref{sec:exp:ablation}), we {always evaluate} on
eLoFTR~\cite{wang2024eloftr} (training mixes eLoFTR and SIFT), per-pair RANSAC-verified{~\cite{fischler1981random}}, with $K_{\text{test}}{=}100$ on every benchmark. {Further} training details are in \cref{app:impl}. Code and trained weights will be released upon publication.

\subsection{Camera Pose Estimation}
\label{sec:exp:pose}

We evaluate camera pose on CO3Dv2 and RealEstate10K (\cref{tab:pose}) following the $10$-frame protocol in VGGT~\cite{vggt2024}. CO3Dv2 covers {object-centric} wide-baseline pairs, while RealEstate10K covers indoor and outdoor video motion. We additionally evaluate pose on four {further} multi-view benchmarks, DTU, Mip-NeRF~360, ETH3D, and IMC2021 (\cref{tab:aggregated_results_all_horizontal}), each scored at the tighter @3\textdegree{}/@5\textdegree{}/@10\textdegree{} thresholds (per-dataset composition in \cref{app:datasets}).

\begin{table}[t]
\centering
\renewcommand{\arraystretch}{0.8}
\caption{\textbf{Camera pose estimation on CO3Dv2 and RealEstate10K}
(10 frames per scene).
\textit{%
\textbf{Bold}/\underline{underline}: best/second per column within each group.
\colorbox{gray!18}{Grey}: image-to-geometry methods.
$^\dagger$: ground-truth intrinsics plus a final classical bundle adjustment. Params: from each checkpoint. Time: seconds per scene on one H100{, as matching $+$ solver}.
$^\ddagger$: same cached eLoFTR correspondences as DeepBA{, given to {COLMAP/GLOMAP/GASfM/RESfM as multi-view tracks} (\cref{app:datasets})}.
}}
\label{tab:pose}
\setlength{\tabcolsep}{4pt}
\small
\resizebox{\columnwidth}{!}{%
\begin{tabular}{l|ccc|c|c|c}
\toprule
\multirow{2}{*}{Method} & \multicolumn{3}{c|}{CO3Dv2} & RealEstate10K & \multirow{2}{*}{Params} & \multirow{2}{*}{Time} \\
 & RRA@15\textdegree{}$\uparrow$ & RTA@15\textdegree{}$\uparrow$ & AUC@30\textdegree{}$\uparrow$ & AUC@30\textdegree{}$\uparrow$ & & \\
\midrule
\multicolumn{7}{l}{\textit{Classical SfM}} \\
    COLMAP {(SIFT)}~\cite{schonberger2016sfm} & \hc{86.3} & \hcu{79.9} & \hcu{74.9} & \hc{46.5} & -- & {$\sim$2$+$3\,s} \\
    GLOMAP {(SIFT)}~\cite{pan2024glomap} & \hcu{86.6} & \hc{76.2} & \hc{71.1} & \hc{71.7} & -- & {$\sim$2$+$2\,s} \\
    COLMAP (eLoFTR)$^\ddagger$~\cite{schonberger2016sfm} & \hcb{89.7} & \hcb{83.5} & \hcb{77.0} & \hcb{80.9} & -- & {$\sim$7$+$11\,s} \\
    GLOMAP (eLoFTR)$^\ddagger$~\cite{pan2024glomap} & \hc{82.1} & \hc{72.3} & \hc{67.6} & \hcu{72.7} & -- & {$\sim$7$+$5\,s} \\
\midrule
\multicolumn{7}{l}{\textcolor{gray}{\textit{Learned image-to-geometry}}} \\
    \gc{PoseDiffusion~\cite{wang2023posediffusion}} & \gc{87.7} & \gc{80.0} & \gc{71.6} & \gc{39.6} & \gc{39.0M} & \gc{$\sim$115\,s} \\
    \gc{{DUSt3R}~\cite{wang2024dust3r}} & \gc{98.3} & \gc{82.1} & \gc{72.9} & \gc{79.9} & \gc{577.8M} & \gc{$\sim$7\,s} \\
    \gc{MASt3R~\cite{mast3r_eccv24}} & \gcu{99.2} & \gcb{92.1} & \gcu{84.7} & \gcb{91.1} & \gc{695.3M} & \gc{$\sim$9\,s} \\
    \gc{VGGSfM~\cite{wang2024vggsfm}} & \gc{94.3} & \gcu{85.8} & \gc{80.5} & \gcu{89.5} & \gc{281.2M} & \gc{$\sim$10\,s} \\
    \gc{VGGT~\cite{vggt2024}} & \gcb{99.3} & \gcb{92.1} & \gcb{85.1} & \gc{85.3} & \gc{$\sim$1.2B} & \gc{$\sim$0.2\,s} \\
\midrule
\multicolumn{7}{l}{\textit{Learning-based SfM from matches}} \\
    RESfM {(SIFT)}$^\dagger$~\cite{khatib2025resfm} & \hc{62.7} & \hc{55.5} & \hc{51.1} & \hc{48.9} & 0.7M & {$\sim$2$+$13\,s} \\
    GASfM {(SIFT)}$^\dagger$~\cite{brynte2024learning} & \hc{79.1} & \hc{73.9} & \hc{68.6} & \hc{75.4} & 145.2M & {$\sim$2$+$0.5\,s} \\
    RESfM (eLoFTR)$^{\dagger,\ddagger}$~\cite{khatib2025resfm} & \hc{61.1} & \hc{54.0} & \hc{48.2} & \hc{63.7} & 0.7M & {$\sim$7$+$13\,s} \\
    GASfM (eLoFTR)$^{\dagger,\ddagger}$~\cite{brynte2024learning} & \hcu{79.9} & \hcu{77.3} & \hcu{71.2} & \hcu{82.8} & 145.2M & {$\sim$7$+$2\,s} \\

    \textbf{DeepBA} & \hcb{97.4} & \hcb{89.2} & \hcb{78.6} & \hcb{84.0} & {2.04M} & {$\sim$7$+$3\,s} \\
\bottomrule
\end{tabular}%
}
\end{table}

\begin{figure}[!t]
    \centering
    \sbox0{\includegraphics[width=0.94\columnwidth]{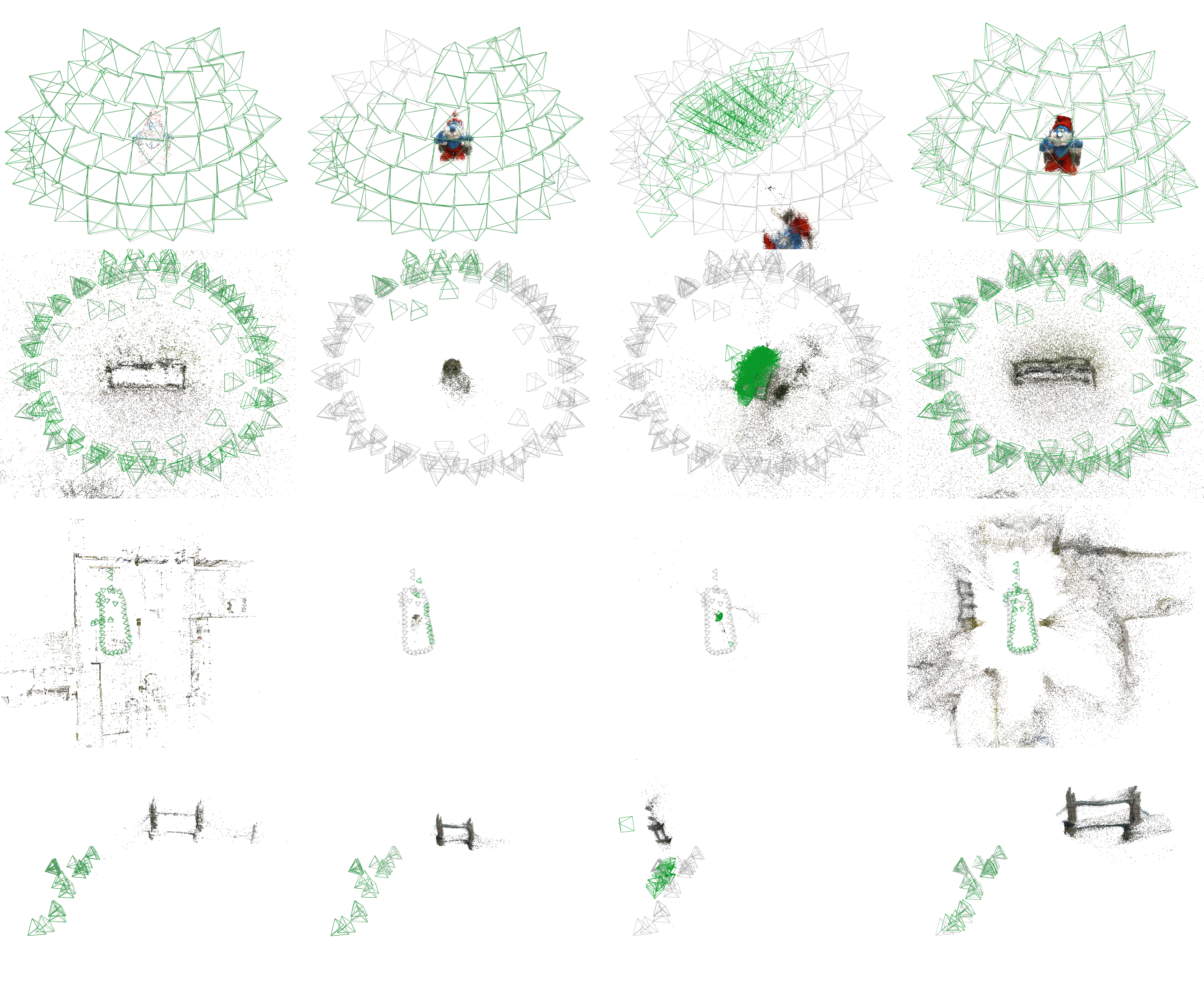}}%
    {\tiny
    \hspace*{1.1em}%
    \makebox[0.235\columnwidth]{COLMAP {(eLoFTR)}}%
    \makebox[0.235\columnwidth]{RESfM {(eLoFTR)}}%
    \makebox[0.235\columnwidth]{GASfM {(eLoFTR)}}%
    \makebox[0.235\columnwidth]{DeepBA}\par\vspace{1pt}}%
    \def\qgrow#1{\parbox[b][\dimexpr\ht0/4\relax][c]{1.1em}{\centering
        \rotatebox{90}{\makebox[0pt][c]{\tiny #1}}}}%
    \leavevmode\vbox{\offinterlineskip
        \hbox{\qgrow{DTU (49)}}%
        \hbox{\qgrow{Mip-NeRF 360 (194)}}%
        \hbox{\qgrow{ETH3D (44)}}%
        \hbox{\qgrow{IMC2021 (25)}}}%
    \usebox0
    \caption{\textbf{Qualitative comparison} (rows: \texttt{scan82},
    \texttt{bicycle}, \texttt{delivery\_area}, and \texttt{london\_bridge}; the number after each dataset is {the number of input images}):
    3D points, estimated cameras (\cgreen{}), GT cameras (\cgrey{}).}
    \label{fig:qual_grid}
    \vspace{-0.4cm}
\end{figure}

\begin{table*}[t]
\centering
\renewcommand{\arraystretch}{0.8}
\caption{\textbf{Multi-view pose accuracy (RRA/RTA) on DTU, Mip-NeRF 360, ETH3D, and IMC2021.}
\textit{%
\textbf{Bold}/\underline{underline}: best/second per column within each group.
\colorbox{gray!18}{Grey}: image-to-geometry methods, for reference.
$^\dagger${, $^\ddagger$}: as in \cref{tab:pose}.%
}}
\label{tab:aggregated_results_all_horizontal}
\begin{adjustbox}{max width=\linewidth}
\begin{tabular}{l | ccc | ccc | ccc | ccc | ccc | ccc | ccc | ccc}
\toprule
\multirow{3}{*}{{Method}} & \multicolumn{6}{c|}{DTU~\cite{Jensen_2014_CVPR} (49 img/scene)} & \multicolumn{6}{c|}{Mip-NeRF 360~\cite{barron2022mipnerf360} (125--311 img/scene)} & \multicolumn{6}{c|}{ETH3D~\cite{schops2017multi} (14--76 img/scene)} & \multicolumn{6}{c}{IMC2021~\cite{jin2021image} (5/10/25-view bags)} \\
\cmidrule(lr){2-7} \cmidrule(lr){8-13} \cmidrule(lr){14-19} \cmidrule(l){20-25}
& \multicolumn{3}{c|}{RRA{$\uparrow$}} & \multicolumn{3}{c|}{RTA{$\uparrow$}} & \multicolumn{3}{c|}{RRA{$\uparrow$}} & \multicolumn{3}{c|}{RTA{$\uparrow$}} & \multicolumn{3}{c|}{RRA{$\uparrow$}} & \multicolumn{3}{c|}{RTA{$\uparrow$}} & \multicolumn{3}{c|}{RRA{$\uparrow$}} & \multicolumn{3}{c}{RTA{$\uparrow$}} \\
& @3\textdegree{} & @5\textdegree{} & @10\textdegree{} & @3\textdegree{} & @5\textdegree{} & @10\textdegree{} & @3\textdegree{} & @5\textdegree{} & @10\textdegree{} & @3\textdegree{} & @5\textdegree{} & @10\textdegree{} & @3\textdegree{} & @5\textdegree{} & @10\textdegree{} & @3\textdegree{} & @5\textdegree{} & @10\textdegree{} & @3\textdegree{} & @5\textdegree{} & @10\textdegree{} & @3\textdegree{} & @5\textdegree{} & @10\textdegree{} \\
\midrule
\multicolumn{25}{l}{\textit{Classical SfM}} \\
    COLMAP {(SIFT)}\cite{schonberger2016sfm} & \hc{87.3} & \hc{87.3} & \hc{87.3} & \hc{65.6} & \hc{74.6} & \hc{82.6} & \hcb{100.0} & \hcb{100.0} & \hcb{100.0} & \hcb{99.9} & \hcb{100.0} & \hcb{100.0} & \hcb{96.7} & \hcb{96.7} & \hcb{96.7} & \hcb{94.9} & \hcb{95.8} & \hcb{96.4} & \hcu{94.0} & \hcu{94.9} & \hc{95.7} & \hc{74.7} & \hc{82.5} & \hc{89.3} \\
    GLOMAP {(SIFT)}\cite{pan2024glomap} & \hc{85.9} & \hc{86.8} & \hc{87.2} & \hc{68.7} & \hc{73.4} & \hc{78.6} & \hcb{100.0} & \hcb{100.0} & \hcb{100.0} & \hcb{99.9} & \hcb{100.0} & \hcb{100.0} & \hcu{94.4} & \hcu{94.5} & \hcu{96.6} & \hcu{86.3} & \hcu{89.3} & \hcu{91.5} & \hc{92.2} & \hc{93.9} & \hc{96.0} & \hc{73.9} & \hc{81.4} & \hc{88.2} \\
    COLMAP (eLoFTR)$^\ddagger$\cite{schonberger2016sfm} & \hcb{100.0} & \hcb{100.0} & \hcb{100.0} & \hcb{91.8} & \hcb{97.1} & \hcb{99.5} & \hcb{100.0} & \hcb{100.0} & \hcb{100.0} & \hc{99.6} & \hc{99.8} & \hcb{100.0} & \hc{76.5} & \hc{76.6} & \hc{76.9} & \hc{69.1} & \hc{72.8} & \hc{75.5} & \hcb{95.8} & \hcb{96.7} & \hcu{97.2} & \hcb{75.4} & \hcb{84.3} & \hcb{91.5} \\
    GLOMAP (eLoFTR)$^\ddagger$\cite{pan2024glomap} & \hcu{97.1} & \hcu{97.1} & \hcu{97.9} & \hcu{89.7} & \hcu{94.8} & \hcu{96.8} & \hcb{100.0} & \hcb{100.0} & \hcb{100.0} & \hcu{99.7} & \hcu{99.9} & \hcb{100.0} & \hc{58.4} & \hc{59.7} & \hc{61.6} & \hc{50.2} & \hc{53.5} & \hc{56.9} & \hcb{95.8} & \hcb{96.7} & \hcb{97.8} & \hcu{74.9} & \hcu{83.6} & \hcu{91.2} \\
\midrule
\multicolumn{25}{l}{\textcolor{gray}{\textit{Learned image-to-geometry}}} \\
\gc{MASt3R-SfM \cite{duisterhof2025mast3rsfm}} & \gcu{99.6} & \gcb{100.0} & \gcb{100.0} & \gcu{92.0} & \gcu{97.7} & \gcu{99.5} & \gcu{68.5} & \gcu{79.9} & \gcu{86.9} & \gcu{60.2} & \gcu{75.2} & \gcu{84.0} & \gcu{67.8} & \gcu{74.0} & \gcu{79.0} & \gcb{63.8} & \gcb{68.8} & \gcu{74.6} & \gcu{94.0} & \gcu{96.2} & \gcu{97.5} & \gcu{62.4} & \gcu{76.3} & \gcu{87.7} \\
\gc{VGGT \cite{vggt2024}} & \gcb{100.0} & \gcb{100.0} & \gcb{100.0} & \gcb{99.3} & \gcb{99.9} & \gcb{100.0} & \gcb{97.6} & \gcb{100.0} & \gcb{100.0} & \gcb{87.1} & \gcb{95.3} & \gcb{98.5} & \gcb{84.1} & \gcb{93.4} & \gcb{96.9} & \gcu{51.4} & \gcu{65.0} & \gcb{79.6} & \gcb{99.8} & \gcb{100.0} & \gcb{100.0} & \gcb{71.1} & \gcb{82.4} & \gcb{91.9} \\
\midrule
\multicolumn{25}{l}{\textit{Learning-based SfM from matches}} \\
    RESfM {(SIFT)}$^\dagger$\cite{khatib2025resfm} & \hc{22.0} & \hc{22.0} & \hc{22.0} & \hc{22.3} & \hc{23.3} & \hc{26.0} & \hc{11.6} & \hc{11.8} & \hc{12.0} & \hc{11.5} & \hc{11.7} & \hc{12.1} & \hc{3.1} & \hc{3.5} & \hc{4.5} & \hc{3.0} & \hc{3.2} & \hc{3.8} & \hc{81.7} & \hc{82.6} & \hc{84.6} & \hcu{76.7} & \hcu{79.0} & \hc{81.2} \\
    GASfM {(SIFT)}$^\dagger$\cite{brynte2024learning} & \hcu{38.3} & \hcu{39.9} & \hcu{40.8} & \hcu{39.1} & \hcu{40.5} & \hcu{44.1} & \hc{32.4} & \hc{35.2} & \hc{41.1} & \hc{31.7} & \hc{35.4} & \hc{44.3} & \hc{8.8} & \hc{9.5} & \hc{12.1} & \hc{8.3} & \hc{9.0} & \hc{10.7} & \hc{74.7} & \hc{80.4} & \hc{86.3} & \hc{56.1} & \hc{63.9} & \hc{75.3} \\
    RESfM (eLoFTR)$^{\dagger,\ddagger}$\cite{khatib2025resfm} & \hc{36.6} & \hc{36.7} & \hc{36.8} & \hc{36.8} & \hc{37.2} & \hc{39.3} & \hc{10.7} & \hc{11.1} & \hc{11.3} & \hc{10.5} & \hc{10.8} & \hc{12.0} & \hc{19.1} & \hc{19.3} & \hc{19.6} & \hc{18.3} & \hc{18.8} & \hc{19.4} & \hc{81.0} & \hc{82.2} & \hc{84.8} & \hc{75.5} & \hc{77.9} & \hc{79.8} \\
    GASfM (eLoFTR)$^{\dagger,\ddagger}$\cite{brynte2024learning} & \hc{36.9} & \hc{37.6} & \hc{38.8} & \hc{37.4} & \hc{38.6} & \hc{42.4} & \hcu{49.6} & \hcu{51.9} & \hcu{54.7} & \hcu{50.8} & \hcu{53.5} & \hcu{57.6} & \hcu{21.6} & \hcu{23.9} & \hcu{27.2} & \hcb{21.0} & \hcu{23.9} & \hcu{29.4} & \hcb{89.6} & \hcu{91.0} & \hcu{92.2} & \hcb{83.1} & \hcb{87.3} & \hcb{90.8} \\

    \textbf{\ours{}} & \hcb{73.6} & \hcb{95.6} & \hcb{100.0} & \hcb{69.5} & \hcb{90.8} & \hcb{99.6} & \hcb{53.3} & \hcb{66.7} & \hcb{73.5} & \hcb{58.6} & \hcb{68.2} & \hcb{86.5} & \hcb{33.8} & \hcb{53.4} & \hcb{69.1} & \hcu{18.4} & \hcb{28.8} & \hcb{49.1} & \hcu{86.9} & \hcb{95.6} & \hcb{99.0} & \hc{49.4} & \hc{68.9} & \hcu{87.3} \\
\bottomrule
\end{tabular}
\end{adjustbox}
\end{table*}

\paragraph{CO3Dv2 and RealEstate10K.}
On {both benchmarks} (\cref{tab:pose}), DeepBA leads the {learned methods from matches}. On RealEstate10K, GASfM with the same eLoFTR correspondences {comes closest (AUC@30{\textdegree{}} 82.8 vs.\ 84.0)}, while {GASfM (SIFT)} scores {75.4}. {Against classical SfM, DeepBA outperforms COLMAP and GLOMAP on every metric of both benchmarks, with both SIFT and the shared eLoFTR input.}

\paragraph{{DTU, Mip-NeRF~360, ETH3D and IMC2021.}}
On {these four} benchmarks (\cref{tab:aggregated_results_all_horizontal}), {DeepBA leads the {learned methods from matches} on most metrics. {GASfM and RESfM, given ground-truth intrinsics, are} ahead only in a few cells, mostly on IMC2021 translation.} Against classical SfM {on the shared eLoFTR input, DeepBA comes close to both mappers at 10\textdegree{} on DTU and IMC2021 but falls behind at the tighter thresholds. On Mip-NeRF~360 the mappers are near the ceiling and DeepBA {trails them} (RRA@10{\textdegree{}} 73.5, RTA@10{\textdegree{}} 86.5). On ETH3D, COLMAP is well ahead (RTA@10{\textdegree{}} 75.5 against 49.1, see \cref{app:eth3d}).} We also provide a qualitative comparison in \cref{fig:qual_grid} {and a failure analysis in \cref{app:failures}}.

\paragraph{Camera localization.}
DeepBA's robustness also enables camera localization (\cref{app:localization}). Reusing the same model with the reference cameras held fixed, DeepBA registers query views against a frozen map from {the mean pose of the map cameras}, with no PnP or external solver.

\subsection{Structure Accuracy}
\label{sec:exp:structure}

\begin{table}[!ht]
\centering
\renewcommand{\arraystretch}{0.8}
\caption{\textbf{Structure accuracy and completeness on DTU} (15 test scans).
Succ.: scans with a non-empty object reconstruction.
\textit{\textbf{Bold}/\underline{underline}: best/second on accuracy and completeness metrics within each group.
$^\dagger${, $^\ddagger$}: as in \cref{tab:pose}.} {\colorbox{gray!18}{Grey}: image-to-geometry methods, for reference.}}
\label{tab:structure}
\setlength{\tabcolsep}{3pt}
\small
\resizebox{\columnwidth}{!}{%
\begin{tabular}{l|ccc|ccc|c}
\toprule
\multirow{2}{*}{Method} & \multicolumn{3}{c|}{Accuracy (\%)$\uparrow$} & \multicolumn{3}{c|}{Completeness (\%)$\uparrow$} & \multirow{2}{*}{Succ.} \\
 & @1\,cm & @2\,cm & @5\,cm & @1\,cm & @2\,cm & @5\,cm & \\
\midrule
\multicolumn{8}{l}{\textit{Classical SfM}} \\
    COLMAP {(SIFT)}~\cite{schonberger2016sfm} & \hcb{94.2} & \hcu{97.6} & \hcu{99.4} & \hcu{80.6} & \hcu{95.2} & \hcu{99.6} & 15/15 \\
    GLOMAP {(SIFT)}~\cite{pan2024glomap} & \hc{43.0} & \hc{49.3} & \hc{65.0} & \hc{30.1} & \hc{40.7} & \hc{51.6} & 11/15 \\
    COLMAP (eLoFTR)$^\ddagger$~\cite{schonberger2016sfm} & \hcu{91.7} & \hcb{98.2} & \hcb{99.8} & \hcb{96.6} & \hcb{99.7} & \hcb{100.0} & {15/15} \\
    GLOMAP (eLoFTR)$^\ddagger$~\cite{pan2024glomap} & \hc{64.6} & \hc{71.3} & \hc{73.1} & \hc{68.1} & \hc{72.5} & \hc{73.3} & 12/15 \\
\midrule
\multicolumn{8}{l}{\textcolor{gray}{\textit{Learned image-to-geometry}}} \\
    \gc{MASt3R-SfM~\cite{duisterhof2025mast3rsfm}} & \gcu{88.9} & \gcu{97.4} & \gcb{99.6} & \gcu{85.5} & \gcu{94.1} & \gcu{98.4} & \gc{15/15} \\
    \gc{VGGT~\cite{vggt2024}} & \gcb{95.4} & \gcb{98.0} & \gcu{99.2} & \gcb{97.2} & \gcb{98.8} & \gcb{99.9} & \gc{15/15} \\
\midrule
\multicolumn{8}{l}{\textit{Learning-based SfM from matches}} \\
    RESfM {(SIFT)}$^\dagger$~\cite{khatib2025resfm} & \hc{9.3} & \hc{21.7} & \hc{38.0} & \hc{0.9} & \hc{3.9} & \hc{17.4} & 10/15 \\
    GASfM {(SIFT)}$^\dagger$~\cite{brynte2024learning} & \hcu{33.0} & \hcu{39.3} & \hc{46.4} & \hc{25.4} & \hc{35.3} & \hcu{47.3} & 13/15 \\
    RESfM (eLoFTR)$^{\dagger,{\ddagger}}$~\cite{khatib2025resfm} & \hc{20.6} & \hc{28.1} & \hcu{52.5} & \hc{19.5} & \hc{27.5} & \hc{46.5} & 11/15 \\
    GASfM (eLoFTR)$^{\dagger,{\ddagger}}$~\cite{brynte2024learning} & \hc{30.9} & \hc{34.5} & \hc{42.5} & \hcu{35.6} & \hcu{41.8} & \hc{47.1} & 14/15 \\
    \textbf{\ours{}} & \hcb{34.1} & \hcb{62.3} & \hcb{98.5} & \hcb{37.1} & \hcb{59.5} & \hcb{96.1} & 15/15 \\

\bottomrule

\end{tabular}%
}
\end{table}

Beyond poses, \cref{tab:structure} evaluates the recovered structure on
DTU against the official structured-light reference scans. {Among the learning-based SfM methods from matches,} \ours{} is the
strongest in every column and the only one
to produce a non-empty object reconstruction on all $15$ scans---even though
GASfM and RESfM use ground-truth intrinsics. The {learned image-to-geometry methods} are near the ceiling on this
object-centric benchmark.

\subsection{Ablations and Analysis}
\label{sec:exp:ablation}

\begin{table}[!ht]
\centering
\setlength{\tabcolsep}{4pt}
\resizebox{\columnwidth}{!}{%
\begin{tabular}{@{}c@{\hspace{2em}}c@{}}
\begin{tabular}[t]{l|ccc}
\toprule
 & \shortstack{RRA\\@15\textdegree{}$\uparrow$} & \shortstack{RTA\\@15\textdegree{}$\uparrow$} & \shortstack{AUC\\@30\textdegree{}$\uparrow$} \\
\midrule
\textbf{Default} & {97.4} & {89.2} & {78.6} \\
\midrule
$d_h$: $128 \to 64$ & \worse{79.2} & \worse{72.6} & \worse{58.9} \\
$\lambda_{\text{3d}}$: $0.1 \to 0.01$ & \worse{81.7} & \worse{76.6} & \worse{64.2} \\
gauge strip: off $\to$ on & \worse{97.1} & \worse{88.8} & \worse{78.3} \\
\midrule
{GT focal {length}} & \better{97.5} & \better{90.9} & \better{81.5} \\
{post classical BA} & \better{98.0} & \better{91.0} & \better{84.6} \\
\bottomrule
\end{tabular}
&
\begin{tabular}[t]{l|ccc}
\toprule
\shortstack{Corrupted\\fraction} & \shortstack{RRA\\@15\textdegree{}$\uparrow$} & \shortstack{RTA\\@15\textdegree{}$\uparrow$} & \shortstack{AUC\\@30\textdegree{}$\uparrow$} \\
\midrule
{\textbf{Default}} & {97.4} & {89.2} & {78.6} \\
\midrule
$5\%$  & 95.5 & {86.8} & {74.5} \\
$10\%$ & {95.7} & {86.4} & {73.4} \\
$50\%$ & {86.8} & {77.5} & {62.4} \\
$90\%$ & {65.2} & {58.6} & {42.7} \\
$100\%$ & {63.4} & {56.3} & {40.7} \\

\bottomrule
\end{tabular}
\end{tabular}%
}

\caption{\textbf{Ablations (left) and outlier robustness (right)} on CO3Dv2 (protocol of \cref{tab:pose}). \emph{Left:} each row changes one factor vs.\ the default, \better{green}/\worse{red} marking gain/loss; the gauge strip removes the global {rigid motion} and rescales each iteration.  \emph{Right:} a fraction of matches is slid along the GT epipolar line (Sampson${\approx}0$).}
\label{tab:ablation_L}
\label{tab:outlier}

\end{table}

\begin{table*}[t]
\centering
\renewcommand{\arraystretch}{0.8}
\caption{\textbf{DeepBA generalizes across upstream matchers} (\cref{sec:exp:ablation}). \textit{\textbf{Bold}/\underline{underline}} and protocols as in \cref{tab:aggregated_results_all_horizontal}.}
\label{tab:matchers}
\begin{adjustbox}{max width=\linewidth}
\begin{tabular}{l | ccc | ccc | ccc | ccc | ccc | ccc | ccc | ccc}
\toprule
\multirow{3}{*}{Matcher} & \multicolumn{6}{c|}{DTU~\cite{Jensen_2014_CVPR} (49 img/scene)} & \multicolumn{6}{c|}{Mip-NeRF 360~\cite{barron2022mipnerf360} (125--311 img/scene)} & \multicolumn{6}{c|}{ETH3D~\cite{schops2017multi} (14--76 img/scene)} & \multicolumn{6}{c}{IMC2021~\cite{jin2021image} (5/10/25-view bags)} \\
\cmidrule(lr){2-7} \cmidrule(lr){8-13} \cmidrule(lr){14-19} \cmidrule(l){20-25}
& \multicolumn{3}{c|}{RRA{$\uparrow$}} & \multicolumn{3}{c|}{RTA{$\uparrow$}} & \multicolumn{3}{c|}{RRA{$\uparrow$}} & \multicolumn{3}{c|}{RTA{$\uparrow$}} & \multicolumn{3}{c|}{RRA{$\uparrow$}} & \multicolumn{3}{c|}{RTA{$\uparrow$}} & \multicolumn{3}{c|}{RRA{$\uparrow$}} & \multicolumn{3}{c}{RTA{$\uparrow$}} \\
& @3\textdegree{} & @5\textdegree{} & @10\textdegree{} & @3\textdegree{} & @5\textdegree{} & @10\textdegree{} & @3\textdegree{} & @5\textdegree{} & @10\textdegree{} & @3\textdegree{} & @5\textdegree{} & @10\textdegree{} & @3\textdegree{} & @5\textdegree{} & @10\textdegree{} & @3\textdegree{} & @5\textdegree{} & @10\textdegree{} & @3\textdegree{} & @5\textdegree{} & @10\textdegree{} & @3\textdegree{} & @5\textdegree{} & @10\textdegree{} \\
\midrule
SIFT~\cite{lowe2004distinctive} & \hc{32.2} & \hc{41.3} & \hc{50.7} & \hc{25.2} & \hc{39.5} & \hc{54.1} & \hc{14.2} & \hc{21.4} & \hc{29.4} & \hc{10.0} & \hc{18.7} & \hc{29.8} & \hc{31.5} & \hc{45.2} & \hc{61.3} & \hc{14.5} & \hc{25.8} & \hc{44.4} & \hc{57.0} & \hc{67.6} & \hc{73.3} & \hc{21.9} & \hc{34.3} & \hc{52.1} \\
SP$+$LG~\cite{detone2018superpoint,lindenberger2023lightglue} & \hcu{69.5} & \hcu{93.4} & \hcb{100.0} & \hc{60.3} & \hc{87.1} & \hc{99.5} & \hcu{64.0} & \hcu{72.2} & \hcu{82.0} & \hcu{64.3} & \hcu{77.7} & \hcb{91.7} & \hcu{35.2} & \hcb{55.7} & \hcb{72.5} & \hcu{21.5} & \hcu{35.9} & \hcu{57.6} & \hcu{87.1} & \hcu{96.1} & \hcu{99.1} & \hc{46.7} & \hc{66.7} & \hc{86.2} \\
ALIKED$+$LG~\cite{zhao2023aliked,lindenberger2023lightglue} & \hc{60.6} & \hc{84.1} & \hc{93.3} & \hc{54.8} & \hc{77.7} & \hc{92.9} & \hcb{79.4} & \hcb{85.0} & \hcb{86.7} & \hcb{79.4} & \hcb{84.7} & \hc{88.8} & \hc{28.3} & \hc{47.1} & \hcu{69.5} & \hc{17.9} & \hc{31.8} & \hc{57.1} & \hcb{87.5} & \hcb{96.2} & \hcb{99.2} & \hc{47.9} & \hc{68.0} & \hc{86.8} \\
    eLoFTR~\cite{wang2024eloftr} & \hcb{73.6} & \hcb{95.6} & \hcb{100.0} & \hcu{69.5} & \hcu{90.8} & \hcu{99.6} & \hc{53.3} & \hc{66.7} & \hc{73.5} & \hc{58.6} & \hc{68.2} & \hc{86.5} & \hc{33.8} & \hcu{53.4} & \hc{69.1} & \hc{18.4} & \hc{28.8} & \hc{49.1} & \hc{86.9} & \hc{95.6} & \hc{99.0} & \hcu{49.4} & \hcu{68.9} & \hcu{87.3} \\
DKM~\cite{edstedt2023dkm} & \hc{65.5} & \hc{92.9} & \hcu{99.7} & \hcb{74.6} & \hcb{95.7} & \hcb{99.7} & \hc{61.3} & \hc{68.9} & \hc{74.5} & \hc{63.0} & \hc{71.3} & \hcu{90.0} & \hcb{36.1} & \hc{52.4} & \hc{66.1} & \hcb{26.3} & \hcb{41.4} & \hcb{59.5} & \hc{86.6} & \hc{95.2} & \hc{98.0} & \hcb{51.2} & \hcb{70.3} & \hcb{87.4} \\
\bottomrule
\end{tabular}
\end{adjustbox}
\end{table*}

\paragraph{{Configuration ablations.}}
\cref{tab:ablation_L} (left) ablates the configuration on CO3Dv2. Hidden width $d_h$ matters most: halving it to $d_h{=}64$ drops AUC@30\textdegree{} from $78.6$ to $58.9$, so we keep $d_h{=}128$. Weakening the {3D consistency} loss to $\lambda_{\text{3d}}{=}0.01$ is also costly ($78.6 \rightarrow 64.2$), indicating the 3D term contributes beyond the pose losses. Enabling the per-iteration gauge strip leaves accuracy almost unchanged, so the gauge freedom needs no explicit fixing as expected. {The last two rows change only the evaluation. Ground-truth focal lengths leave RRA almost unchanged but raise RTA, as they remove the focal--depth ambiguity (\cref{app:focaldepth}). A classical BA, started from DeepBA's output, raises all three metrics. {This gain} shows that DeepBA behaves similarly to BA but not identically, which leaves room for improvement.}

\paragraph{Outlier robustness.}
A RANSAC-style geometric filter cannot remove wrong matches that lie on the true epipolar line (Sampson ${\approx}\,0$, e.g., from repetitive structure). We probe this regime on CO3Dv2 scenes (protocol of \cref{tab:pose}), sampling a growing fraction of the matches uniformly across all pairs of a scene and sliding the destination endpoint of each {by} $0$--$50\%$ of the image width along its ground-truth epipolar line. \cref{tab:outlier} (right) shows slow
degradation: {against the clean run on the same CO3Dv2 scenes ($78.6$)}, $10\%$ corruption costs {$5.2$} AUC@30\textdegree{} points and $50\%$ costs {$16.2$}.
{Even at $100\%$, accuracy stays at $40.7$} rather than collapsing, because a match slid along its own
epipolar line still constrains that pair's rotation and translation direction.

\paragraph{Upstream matcher.}
\looseness=-1
DeepBA consumes only 2D matches, so the same trained network can run on any front-end without retraining. We evaluate it with five front-ends---sparse SIFT{~\cite{lowe2004distinctive}}, the sparse learned SuperPoint$+$LightGlue {(SP$+$LG)}~\cite{detone2018superpoint,lindenberger2023lightglue} and ALIKED$+$LightGlue {(ALIKED$+$LG)}~\cite{zhao2023aliked,lindenberger2023lightglue}, semi-dense eLoFTR~\cite{wang2024eloftr}, and dense DKM~\cite{edstedt2023dkm}---on the four multi-view benchmarks (\cref{tab:matchers}). The four learned
front-ends {do not rank by match density}, so {density} alone does not drive accuracy. {SIFT gives the lowest accuracy on most of the metrics.}

\begin{figure}[t]
  \centering
  \includegraphics[width=\linewidth]{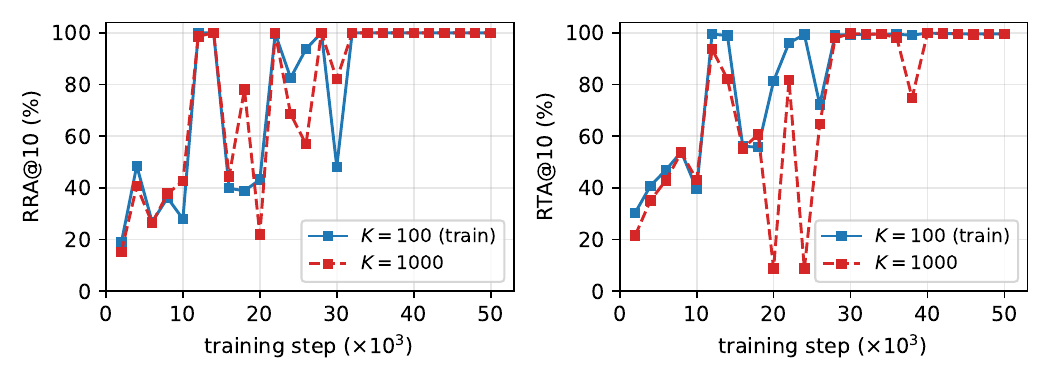}
  \caption{\textbf{Generalizing beyond the training iteration budget}
  (RRA@10\textdegree{} left, RTA@10\textdegree{} right, on \texttt{bicycle}). {Each training checkpoint is evaluated at
  $K{=}100$ (\textcolor[HTML]{1F77B4}{blue}) and $K{=}1000$ (\textcolor[HTML]{D62728}{red}).}}
  \label{fig:rta_step}
\end{figure}

\paragraph{Generalization beyond the training budget.}
We train \ours{} with $K_{\text{train}}{=}100$ refinement iterations on view graphs of at most $48$ cameras, yet at test time it {extends} beyond both: to more iterations and to larger scenes. On iterations, \cref{fig:rta_step} runs $K{=}1000$ ($10\times$ the training budget) across checkpoints: early extrapolation is metastable and can \emph{degrade} the estimate, but stabilizes by {${\sim}40$k} steps. On size, \ours{} handles Mip-NeRF~360 scenes of up to {${\sim}311$} cameras ({${\sim}6.5\times$} its largest training graph).

\begin{figure}[t]
  \centering
  \includegraphics[width=\linewidth]{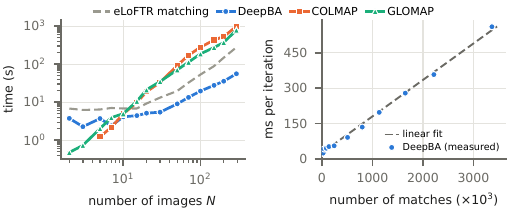}
  \caption{{\emph{Left:} time after matching on the first $N$ images of Mip-NeRF~360 \texttt{bonsai}, with every method using the same eLoFTR matches on one H100 (matching time dashed). \emph{Right:} time per iteration vs.\ number of matches.}}
  \label{fig:runtime_n}
\end{figure}

\paragraph{Runtime scaling.}
{\cref{fig:runtime_n} times {DeepBA, COLMAP and GLOMAP} on the same eLoFTR matches of the first $N$ images of \texttt{bonsai}, so the methods differ only in the solver (matching: $279$\,s at $N{=}292$). From $N{=}10$ on, \ours{} is the fastest {of the three}. At $N{=}292$ ($3.4$M matches), its $K{=}100$ iterations take {$56$}\,s with a GPU memory peak of $24$\,GB, against $988$\,s ($0.9$\,GB) for COLMAP and $793$\,s ($0.5$\,GB) for GLOMAP with their GPU solvers. \ours{}'s per-iteration cost is linear in the number of matches, {$25$}\,ms $+$ {$155$}\,ns per match.}

\section{Conclusion}
We presented DeepBA, a lightweight graph neural network that amortizes bundle
adjustment. From the 2D matches of any off-the-shelf matcher and a trivial cold
start, it jointly infers camera poses, intrinsics, and scene structure for arbitrarily many views.
Across six benchmarks, DeepBA outperforms prior learned SfM from matches
{on most metrics} with {${\sim}71\times$} fewer parameters than GASfM, the strongest of them, and approaches
image-to-geometry transformers that consume the images.
{{Four} limits remain: the focal--depth ambiguity on planar scenes, {a drift of the learned update on some scenes,} the accuracy at tight thresholds, and the pinhole camera model.}
We hope DeepBA encourages further work on learned solvers for multi-view
geometry.

{
    \small
    \bibliographystyle{ieeenat_abbrev}
    \bibliography{main}
}

\clearpage
\appendix
\section{{Method Details}}
\label{app:method-details}

\subsection{{Lie-Group and Pose Conventions}}
\label{app:conventions}

We use $\mathrm{Log}: \mathrm{SE}(3) \rightarrow \mathbb{R}^6$ and
$\mathrm{Exp}: \mathbb{R}^6 \rightarrow \mathrm{SE}(3)$ for the standard
Lie-group logarithm and exponential between a $4{\times}4$ rigid-body pose
matrix and its $6$-dim tangent vector. We overload $\mathrm{Log}$
for $\mathrm{SO}(3)$: $\mathrm{Log}: \mathrm{SO}(3) \to \mathbb{R}^3$ and  $R: \mathbb{R}^3 \rightarrow \mathrm{SO}(3)$. $T_i=T_{iw}$ maps world coordinates to camera-$i$ coordinates,
$\mathbf{x}_i = R_i \mathbf{x}_w + \mathbf{t}_i$; hence
$T^{\text{abs}}_{ij} = T^{\text{abs}}_i (T^{\text{abs}}_j)^{-1}$ maps
camera-$j$ to camera-$i$ coordinates.

\paragraph{Component-wise state update ($\boxplus$, \cref{eq:boxplus}).}
The absolute pose composes on the group, $T^{\text{abs}}_i \leftarrow \mathrm{Exp}(\Delta T^{\text{abs}}_i)\,T^{\text{abs}}_i$, and the log-FOV and log-depth maps update additively, $\phi_i \leftarrow \phi_i + \Delta\phi_i$, $D_i \leftarrow D_i + \Delta D_i$. The decoupled per-edge relative pose $(\hat{\mathbf{t}}^{\text{rel}}_{ij}\in S^2,\ \boldsymbol{\omega}^{\text{rel}}_{ij}\in\mathbb{R}^3)$ of \cref{sec:method:problem} is updated component-wise. Writing $\Delta T^{\text{rel}}_{ij} = (\Delta\mathbf{t}_{ij}, \Delta\boldsymbol{\omega}_{ij})$, the rotation is composed on the group and the translation direction is added and renormalized onto $S^2$:
\begin{equation}
    \boldsymbol{\omega}^{\text{rel}}_{ij} \leftarrow \mathrm{Log}\!\bigl(R(\Delta\boldsymbol{\omega}_{ij})\, R(\boldsymbol{\omega}^{\text{rel}}_{ij})\bigr), \qquad
    \hat{\mathbf{t}}^{\text{rel}}_{ij} \leftarrow \frac{\hat{\mathbf{t}}^{\text{rel}}_{ij}+\Delta\mathbf{t}_{ij}}{\bigl\|\hat{\mathbf{t}}^{\text{rel}}_{ij}+\Delta\mathbf{t}_{ij}\bigr\|},
\end{equation}

\subsection{{Focal Length} $\leftrightarrow$ FOV}
\label{app:intrinsics}
Writing $\mathrm{FOV}_i$ for camera $i$'s field-of-view angle in radians, the
log-FOV is $\phi_i := \log \mathrm{FOV}_i$; the simple-pinhole intrinsic is
\begin{equation}
    K_i = K(\phi_i) = \begin{pmatrix} f_i & 0 & 0.5 \\ 0 & {a_i} f_i & 0.5 \\ 0 & 0 & 1 \end{pmatrix}, \quad f_i = \frac{1}{2\tan\!\bigl(\exp(\phi_i)/2\bigr)},
    \label{eq:intrinsics}
\end{equation}
with image coordinates normalized to $[0,1]^2$ {per axis and $a_i = W_i/H_i$ the aspect ratio of image $i$, so $\mathrm{FOV}_i$ is the horizontal field of view and pixels stay square}.

\subsection{{Pose-Graph Residual}}
\label{app:pose-graph}

The per-edge pose-graph residual (\cref{sec:method:features:pair}) measures the
disagreement between the directly predicted relative pose $T^{\text{rel}}_{ji}$ and
the one derived from the absolute poses, $T^{\text{abs}}_{ji}$. The rotation term
$\boldsymbol{\delta}^{\text{R}}_{ji}$ is that disagreement in the tangent space,
\begin{equation}
    \boldsymbol{\delta}^{\text{R}}_{ji} = \mathrm{Log}\!\bigl(R^{\text{rel}}_{ji}\,(R^{\text{abs}}_{ji})^{\top}\bigr).
    \label{eq:rot-residual}
\end{equation}
The translation term compares directions only. Write
$\hat{\mathbf{t}}^{\text{abs}}_{ji}$ for the unit-normalized translation of
$T^{\text{abs}}_{ji}$. Then $\boldsymbol{\delta}^{\text{t}}_{ji}$ is the minimal
rotation carrying $\hat{\mathbf{t}}^{\text{abs}}_{ji}$ onto
$\hat{\mathbf{t}}^{\text{rel}}_{ji}$, written as an axis--angle $3$-vector
\begin{equation}
    \boldsymbol{\delta}^{\text{t}}_{ji} = \theta\,\hat{\mathbf{n}}, \qquad
    \theta = \angle\bigl(\hat{\mathbf{t}}^{\text{rel}}_{ji},\, \hat{\mathbf{t}}^{\text{abs}}_{ji}\bigr), \qquad
    \hat{\mathbf{n}} \propto \hat{\mathbf{t}}^{\text{abs}}_{ji} \times \hat{\mathbf{t}}^{\text{rel}}_{ji}.
    \label{eq:tdir-residual}
\end{equation}

\subsection{{Match-Feature Residuals}}
\label{app:match-feats}

Throughout this subsection $e$ is a match on the pair
$(i,j)\in\mathcal{E}_{\text{pair}}$, with endpoints $\mathbf{u}^e_{i}$ in image $i$
and $\mathbf{u}^e_{j}$ in image $j$, written $\mathbf{u}^e_{\text{src}}$ and
$\mathbf{u}^e_{\text{dst}}$ in \cref{sec:method:features:match}. Both are in the
normalized image coordinates of \cref{eq:intrinsics}, and $\tilde{\mathbf{u}} =
(\mathbf{u}^{\top},1)^{\top}$ denotes the homogeneous lift of a 2D point.

The four match-feature quantities of \cref{sec:method:features:match} are all
closed form. The \emph{epipolar vector} is computed from the directly predicted
relative pose. Its translation magnitude is irrelevant here, since the epipolar
geometry does not depend on it and $\hat{\mathbf{t}}^{\text{rel}}_{ji}$ is already a
unit direction. The
essential matrix
$E_{ji} = [\hat{\mathbf{t}}^{\text{rel}}_{ji}]_\times R^{\text{rel}}_{ji}$, with $R^{\text{rel}}_{ji} := R(\boldsymbol{\omega}^{\text{rel}}_{ji})$, defines
the epipolar line for $\mathbf{u}^e_{i}$ in image $j$ (in normalized
camera coordinates) as
$\mathbf{l}_{ji} = E_{ji}\, K_i^{-1} \tilde{\mathbf{u}}^e_{i}$.
Because the simple pinhole model has isotropic focal length $f_j$, mapping the
point-to-line offset back to image-plane coordinates collapses to a single
scalar, and the shortest image-plane vector from $\mathbf{u}^e_{j}$
to the line is
\begin{equation}
    \boldsymbol{\epsilon}_{e} \;=\; -\, f_j \,\frac{\mathbf{l}_{ji,1:2} \,\bigl(\mathbf{l}_{ji}^\top K_j^{-1} \tilde{\mathbf{u}}^e_{j}\bigr)}{\|\mathbf{l}_{ji,1:2}\|^2},
\end{equation}
where $\mathbf{l}_{ji,1:2}$ is the first two components of $\mathbf{l}_{ji}$.

The remaining reprojection and {3D consistency} residuals use the absolute-pose-derived relative pose $T^{\text{abs}}_{ji}$, whose rotation and translation are $R^{\text{abs}}_{ji}$ and $\mathbf{t}^{\text{abs}}_{ji}$ in the sense of \cref{app:conventions}, so that $\mathbf{x}_j = R^{\text{abs}}_{ji}\mathbf{x}_i + \mathbf{t}^{\text{abs}}_{ji}$. Per-pixel depths are sampled from the current log-depth maps,
\begin{equation}
    d^e_{i} = \exp\!\bigl(D_i[\mathbf{u}^e_{i}]\bigr), \qquad
    d^e_{j} = \exp\!\bigl(D_j[\mathbf{u}^e_{j}]\bigr),
\end{equation}
with $D_i[\mathbf{u}]$ bilinear sampling of the log-depth map at image coordinate $\mathbf{u}$; unprojection gives the per-camera 3D points
\begin{equation}
    \mathbf{P}^e_{i} = d^e_{i}\, K_i^{-1}\tilde{\mathbf{u}}^e_{i}, \qquad
    \mathbf{P}^e_{j} = d^e_{j}\, K_j^{-1}\tilde{\mathbf{u}}^e_{j};
\end{equation}
and the residuals are
\begin{equation}
\begin{aligned}
    \boldsymbol{\rho}_{e} &= \pi\!\bigl(K_j(R^{\text{abs}}_{ji}\mathbf{P}^e_{i}+\mathbf{t}^{\text{abs}}_{ji})\bigr) - \mathbf{u}^e_{j}, \\
    \boldsymbol{\delta}^{\text{3D}}_{e} &= \mathbf{P}^e_{j} - \bigl(R^{\text{abs}}_{ji}\mathbf{P}^e_{i}+\mathbf{t}^{\text{abs}}_{ji}\bigr).
\end{aligned}
\end{equation}
where $\pi\bigl([x,y,z]^{\top}\bigr) = [x/z,\,y/z]^{\top}$ is the perspective
division.

\subsection{{Depth Head}}
\label{app:depth-head}

For each match $e$ on pair $(i,j)$, the depth head scatters
$\mathbf{h}^{\text{match}}_{{e_{ij}}}$ onto camera $j$'s $H_d{\times}W_d$ grid at
$\mathbf{u}^{e}_{\text{dst}}$ (and $\mathbf{h}^{\text{match}}_{{e_{ji}}}$ onto
camera $i$'s grid at $\mathbf{u}^{e}_{\text{src}}$) via the bilinear adjoint of
\texttt{grid\_sample}. The scatter also accumulates a per-cell count $c$, the
bilinear weight of the matches landing in that cell, which divides the summed
features into a per-cell mean. That mean, channel-wise layer-normalized and
concatenated with $\log(1{+}c)$, is
decoded by three $3{\times}3$ convolutions into the log-depth delta
$\Delta D_i$.

\section{{Metric Definitions}}
\label{app:metrics}

\subsection{{Pose Accuracy}}
All pose metrics are computed over the \emph{complete} pair set of each scene. Let $\mathcal{V}$ denote the $N$ evaluation cameras of a scene and $\mathcal{P} = \{(i,j) : 1 \le i < j \le N\}$ its $\binom{N}{2}$ unordered pairs. For a pair $(i,j)$, $e^{ij}_R$ is the geodesic angle between the predicted and ground-truth relative rotations and $e^{ij}_t$ the angle between the predicted and ground-truth relative translation \emph{directions}. Both are measured in degrees, as is every threshold $\tau$ in this subsection. A method that produces no pose for camera $i$ (unregistered, dropped, or non-finite) is assigned $e^{ij}_R = e^{ij}_t = \infty$ for every pair containing $i$ (zero credit). With $\mathbb{I}[\cdot]$ the indicator function ($1$ if the bracketed condition holds, else $0$):
\begin{equation}
\begin{aligned}
    \text{RRA@}\tau &= \frac{100}{|\mathcal{P}|} \sum_{(i,j) \in \mathcal{P}} \mathbb{I}\bigl[e^{ij}_R < \tau\bigr], \\
    \text{RTA@}\tau &= \frac{100}{|\mathcal{P}|} \sum_{(i,j) \in \mathcal{P}} \mathbb{I}\bigl[e^{ij}_t < \tau\bigr],
\end{aligned}
    \label{eq:rra-protocol}
\end{equation}
\begin{equation}
\begin{aligned}
    \text{AUC@30\textdegree{}} &= \frac{1}{30} \int_0^{30} A(\tau)\, d\tau, \\
    A(\tau) &= \frac{100}{|\mathcal{P}|} \,\Bigl|\bigl\{(i,j) \in \mathcal{P} : \max(e^{ij}_R, e^{ij}_t) < \tau\bigr\}\Bigr|.
\end{aligned}
    \label{eq:auc-protocol}
\end{equation}

\subsection{{Structure Accuracy}}

Protocol for \cref{tab:structure}. Each reconstruction is aligned to the DTU world frame (millimeters) by a single similarity transform from its predicted camera centers. Both prediction and reference are restricted to the object region by DTU's per-view masks. Writing $d(\mathbf{x},\mathcal{S})=\min_{\mathbf{y}\in\mathcal{S}}\|\mathbf{x}-\mathbf{y}\|$ for the nearest-neighbor distance and $\mathcal{X},\mathcal{Y}$ for the predicted and reference point sets,
\begin{equation}
\begin{aligned}
\text{Accuracy@}\tau &= \frac{100}{|\mathcal{X}|} \sum_{\mathbf{x}\in\mathcal{X}} \mathbb{I}\bigl[d(\mathbf{x},\mathcal{Y})<\tau\bigr], \\
\text{Completeness@}\tau &= \frac{100}{|\mathcal{Y}|} \sum_{\mathbf{y}\in\mathcal{Y}} \mathbb{I}\bigl[d(\mathbf{y},\mathcal{X})<\tau\bigr].
\end{aligned}
\end{equation}

\section{{Benchmark Composition}}
\label{app:datasets}

We detail the composition of the six evaluation benchmarks introduced in
\cref{sec:experiments}.
\begin{itemize}[leftmargin=*,itemsep=0.2em]
    \item \textbf{CO3Dv2}~\cite{reizenstein21co3d}: object-centric turntable-style videos with wide-baseline pairs, evaluated on 202 held-out scenes across 41 categories, disjoint from our training split. Following the $10$-frame protocol of VGGT~\cite{vggt2024}, we sample 10 frames per scene from a fixed per-scene frame list shared across all methods, and report relative pose accuracy.
    \item \textbf{RealEstate10K}~\cite{zhou2018stereo}: indoor and outdoor video clips with smooth camera motion. We follow the same $10$-frame VGGT protocol~\cite{vggt2024} over 108 clips, drawing the frames from fixed per-clip lists shared across all methods.
    \item \textbf{ETH3D}~\cite{schops2017multi}: the 13 high-resolution scenes of the multi-view training split, with laser-scanned ground truth. No ETH3D image enters our training mixture.
    \item \textbf{DTU}~\cite{Jensen_2014_CVPR}: 15 of the 128 laboratory-captured object scans (49 views each) with {structured-light} ground truth, the subset used by SPARF~\cite{truong2022sparf}.
    \item \textbf{Mip-NeRF 360}~\cite{barron2022mipnerf360}: 7 publicly released indoor and outdoor scenes ($125$--$311$ cameras each) captured in 360\textdegree{} sweeps around a central area.
    \item \textbf{IMC2021}~\cite{jin2021image}: all 9 in-the-wild phototourism scenes of the
    2021 Image Matching Challenge, evaluated under the official bag protocol of $175$
    bags per scene ($100$ of $5$ views, $50$ of $10$ and $25$ of $25$, $1{,}575$ bags in
    total).
\end{itemize}

\paragraph{{Baselines on eLoFTR.}}
{The eLoFTR rows of COLMAP/GLOMAP/RESfM/GASfM receive DeepBA's cached, RANSAC-verified eLoFTR correspondences. COLMAP/GLOMAP/RESfM/GASfM need keypoints shared across pairs. Each image's match endpoints are therefore snapped to square cells, with one keypoint per cell at the mean of its endpoints. Each mapper then runs its own geometric verification and track building. Union-find on a grid cell of $1/400$ of the image width ($4$\,px at $1{,}600$\,px) chains the matches into tracks, which are kept when seen in at least 3 views.}

\section{{Training and Implementation Details}}
\label{app:impl}

These expand the headline configuration of \cref{sec:exp:setup}.

\paragraph{Gradient-iteration sampling.}
Each step runs a {$K_{\text{train}} = 100$}-iteration Markov trajectory but takes gradients on only {the} first 12 iterations plus another 12 iterations sampled uniformly from $[12, 100)$ (24 iterations in total with {backpropagation} of \cref{eq:final-loss}; the remaining 76 run under \texttt{no\_grad}). {We denote this set of 24 iterations by $\mathcal{K}_{\text{grad}}$.} We {backpropagate} \emph{per refinement iteration} and detach the state at each gradient boundary, giving memory $O(1)$ in the number of refinement iterations.

\paragraph{Scene and graph sampling.}
At each training step, we sample a connected set of cameras from a randomly chosen scene, primarily with a Sampson-gated graph walk over wide-baseline pairs. We add random edges {of the view graph} so that every camera shares a view with at least one other.

\paragraph{{Robust kernel.}}
The smooth-$L_1$ (Huber) kernel of \cref{sec:method:training} is
\begin{equation}
    \huber{r} \;=\;
    \begin{cases}
        \tfrac{1}{2}\,r^{2}/\beta, & |r| < \beta,\\[2pt]
        |r| - \tfrac{1}{2}\beta,    & \text{otherwise.}
    \end{cases}
\end{equation}

\section{{Convergence against the Refinement Iteration}}
\label{app:convergence}

\begin{figure}[!ht]
  \centering
  \includegraphics[width=\linewidth]{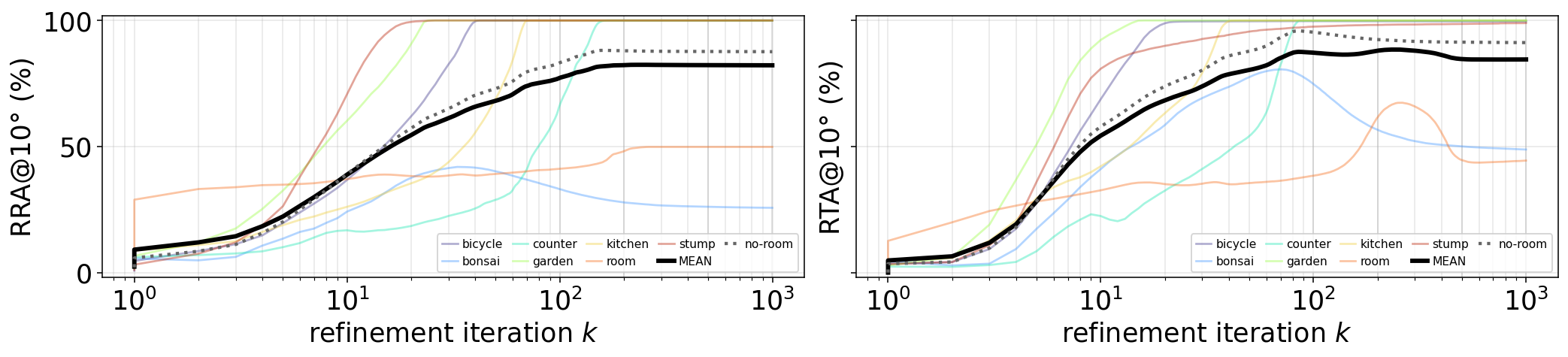}
  \caption{\textbf{Pose accuracy {vs.} refinement iteration} (RRA@10\textdegree{} left, RTA@10\textdegree{}
  right). Light curves are per-scene; the \textbf{solid black} line is the
  across-scene mean, the dotted line the mean excluding the
  view-graph-bottlenecked \texttt{room} scene.}
  \label{fig:rra_iter}
\end{figure}

\cref{fig:rra_iter} profiles accuracy against the number of refinement iterations. We measure it on Mip-NeRF~360, which is absent from the training mixture. Its scenes carry $125$--$311$ views, $2.6\times$ to $6.5\times$ the {largest ($48$-camera)} view graphs \ours{} trains on. From a cold start, accuracy climbs steeply over the first tens of iterations and then saturates, at a scene-dependent rate, with \texttt{garden} and \texttt{stump} settling within ${\sim}30$ iterations and the remaining scenes flattening out well before the end of the profile. We train \ours{} with {$K_{\text{train}}{=}100$}. {The mean RRA@10\textdegree{} keeps rising until ${\sim}150$ iterations and is flat beyond. Extending inference to $K{=}1000$ raises it from $73.5$ to $82.2$, mainly through \texttt{counter}, but lowers the mean RTA@10\textdegree{} from $86.5$ to $84.5$, mainly through \texttt{bonsai}. With no consistent gain from more iterations, we} evaluate every benchmark at $K_{\text{test}}{=}100$.

\section{{{Failure Analysis}}}
\label{app:failures}

\subsection{{ETH3D}}
\label{app:eth3d}

\begin{figure}[!ht]
    \centering
    \includegraphics[width=\linewidth]{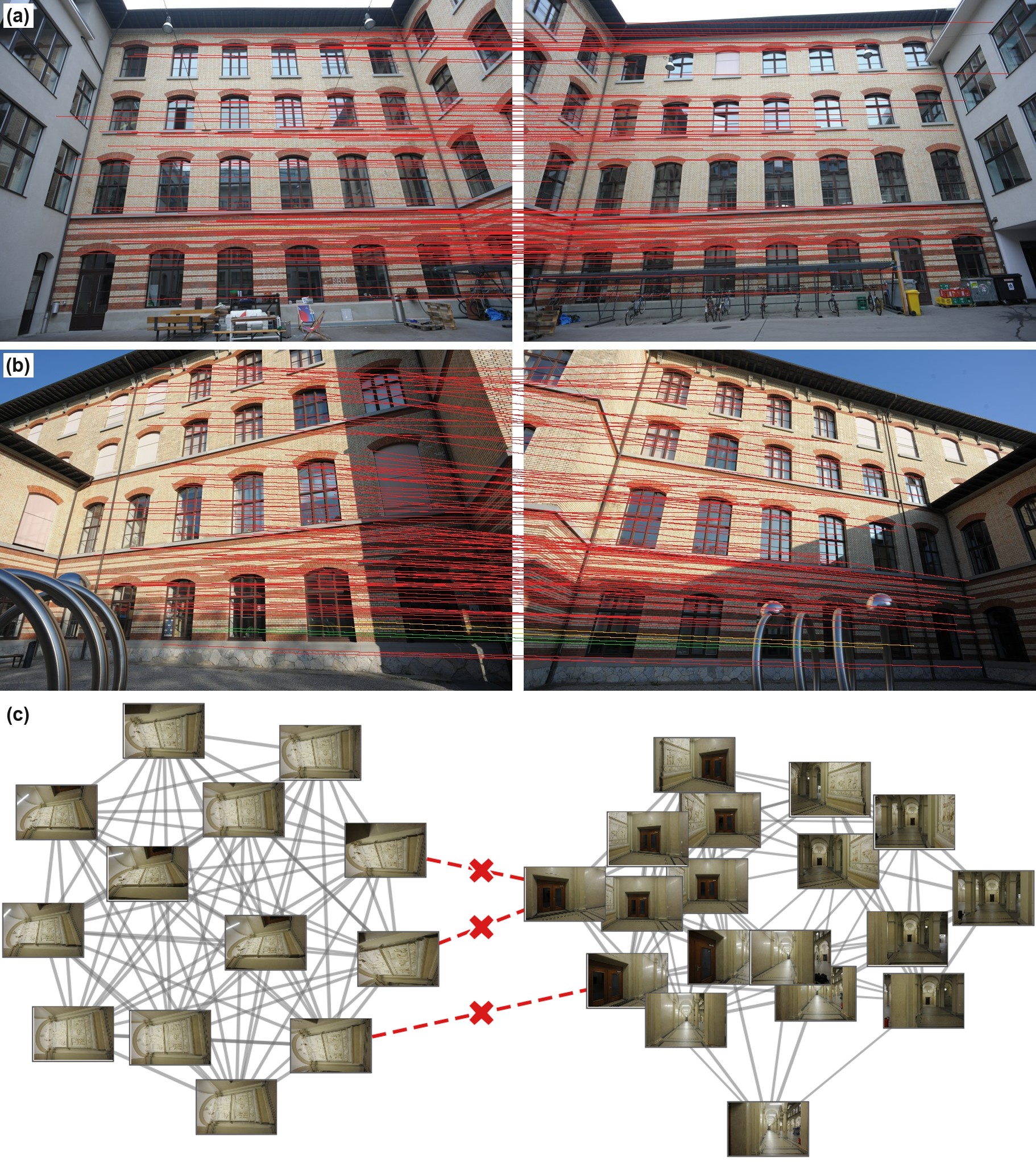}
    \caption{{\textbf{Causes of the ETH3D losses.} (a) \texttt{courtyard} and (b) \texttt{facade}: eLoFTR matches between two different walls with identical windows (red: {ground-truth Sampson error above $20$\,px, green: at most $4$\,px, yellow: in between}). (c) \texttt{relief}: the eLoFTR view graph, with an image per node and an edge per matched pair, splits into the rotated close-ups (left) and the upright views (right), with no matched pair between them.}}
    \label{fig:eth3d_failures}
\end{figure}

\begin{figure}[!ht]
    \centering
    \includegraphics[width=\linewidth]{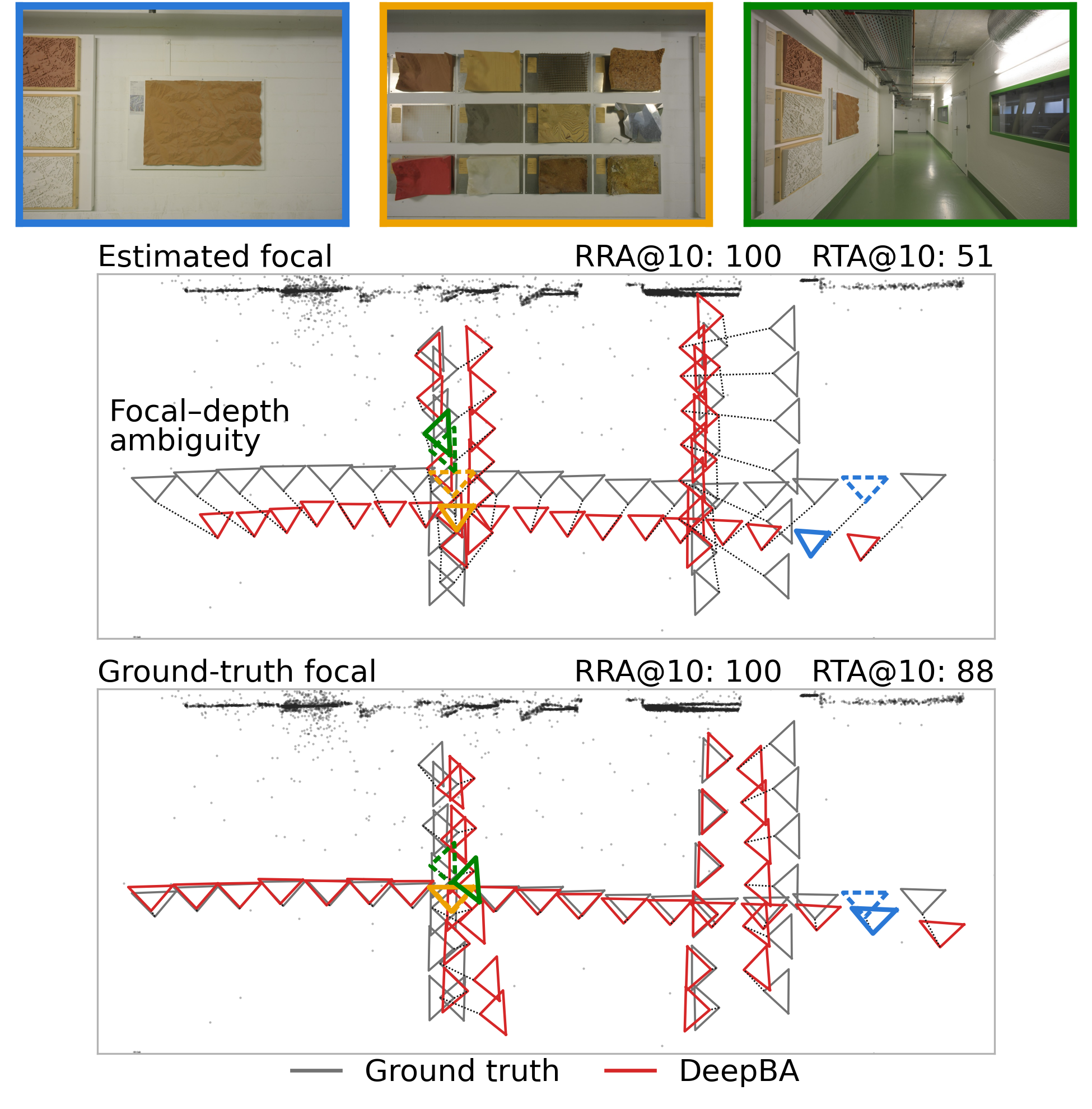}
    \caption{{\textbf{The focal--depth ambiguity on ETH3D \texttt{terrains}.} Top: two views facing the wall {(blue, orange)} and one corridor view {(green)}. Below: \ours{} in \textcolor[HTML]{D62728}{red} against ground truth in \cgrey{}{, with the estimated (upper) and ground-truth (lower) focal length. The three example views keep their colors, solid for \ours{} and dashed for ground truth}.}}
    \label{fig:focal_depth_eth3d}
\end{figure}

{ETH3D is \ours{}'s weakest multi-view benchmark (\cref{tab:aggregated_results_all_horizontal}). Three causes explain most of the losses (\cref{fig:eth3d_failures}{, \cref{fig:focal_depth_eth3d}}).

\emph{Repeated structure.} \texttt{courtyard} and \texttt{facade} show building walls with identical window grids. eLoFTR matches the windows of one wall to those of another, so many pairs consist almost entirely of wrong matches that are mutually consistent (\cref{fig:eth3d_failures}a,b). \ours{} fits these pairs and places the cameras in a wrong configuration. 

\emph{Disconnected views.} Several scenes mix images rotated by $90${\textdegree{}} in the image plane with upright ones (\cref{fig:eth3d_failures}c), or include views of the sky or a ceiling. eLoFTR is not rotation invariant and finds no reliable matches between such groups. The view graph then splits (\cref{fig:eth3d_failures}c).

\emph{Focal--depth ambiguity.} Of the $42$ views of \texttt{terrains}, $18$ face a flat wall head-on. Their rotations are right, but \ours{} overestimates their focal length and places them too far from the wall (\cref{fig:focal_depth_eth3d}), the pattern of \cref{app:focaldepth}. Fixing the focal length to its ground-truth value repairs most of \texttt{terrains}.}

\subsection{{IMC2021}}
\label{app:focaldepth}

\begin{figure}[!ht]
    \centering
    \includegraphics[width=\linewidth]{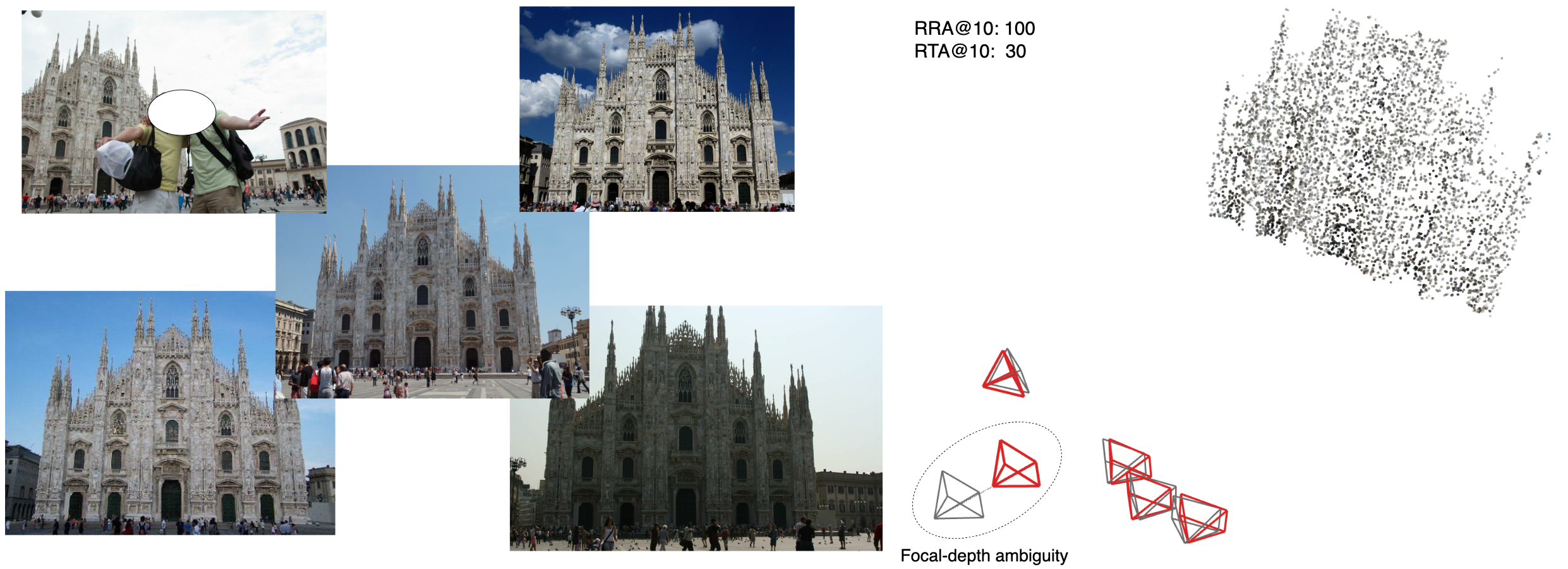}
    \caption{\textbf{The focal--depth ambiguity on one IMC2021 bag}
    (\texttt{milan\_cathedral}, 5 views). \ours{} in \textcolor[HTML]{D62728}{red} against ground truth
    in \cgrey{}; {the circled camera, shown with its ground truth, is displaced by the focal--depth ambiguity}.}
    \label{fig:focal_depth}
\end{figure}

IMC2021 is
the only benchmark in the evaluation set where each camera has different intrinsics. A scene with largely planar facades constrains only the ratio of focal length
to depth rather than either quantity alone. A mis-estimated focal {length} is therefore
absorbed by a compensating depth for that camera, displacing its {center} along
its own optical axis while leaving its viewing directions unchanged
(\cref{fig:focal_depth}). The
relative translation directions tilt while the relative rotations stay intact,
which is the pattern of \cref{tab:aggregated_results_all_horizontal}.

\subsection{{Mip-NeRF~360}}

\begin{figure}[!h]
    \centering
    \begin{subfigure}[t]{0.481\linewidth}
        \centering
        \includegraphics[width=\linewidth]{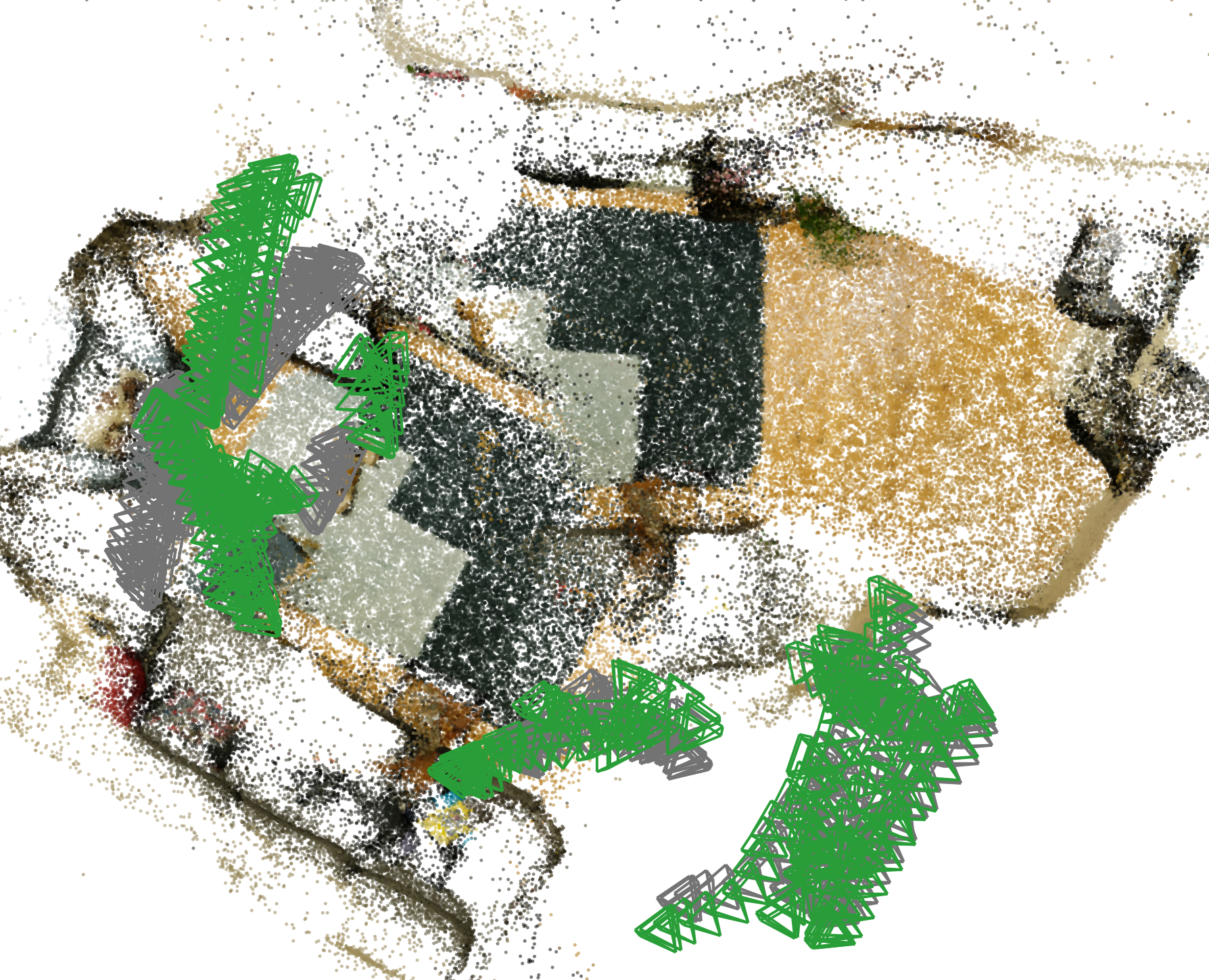}
        \caption{{\texttt{room}: view-graph bottleneck}}
        \label{fig:failure_room}
    \end{subfigure}\hfill
    \begin{subfigure}[t]{0.512\linewidth}
        \centering
        \includegraphics[width=\linewidth]{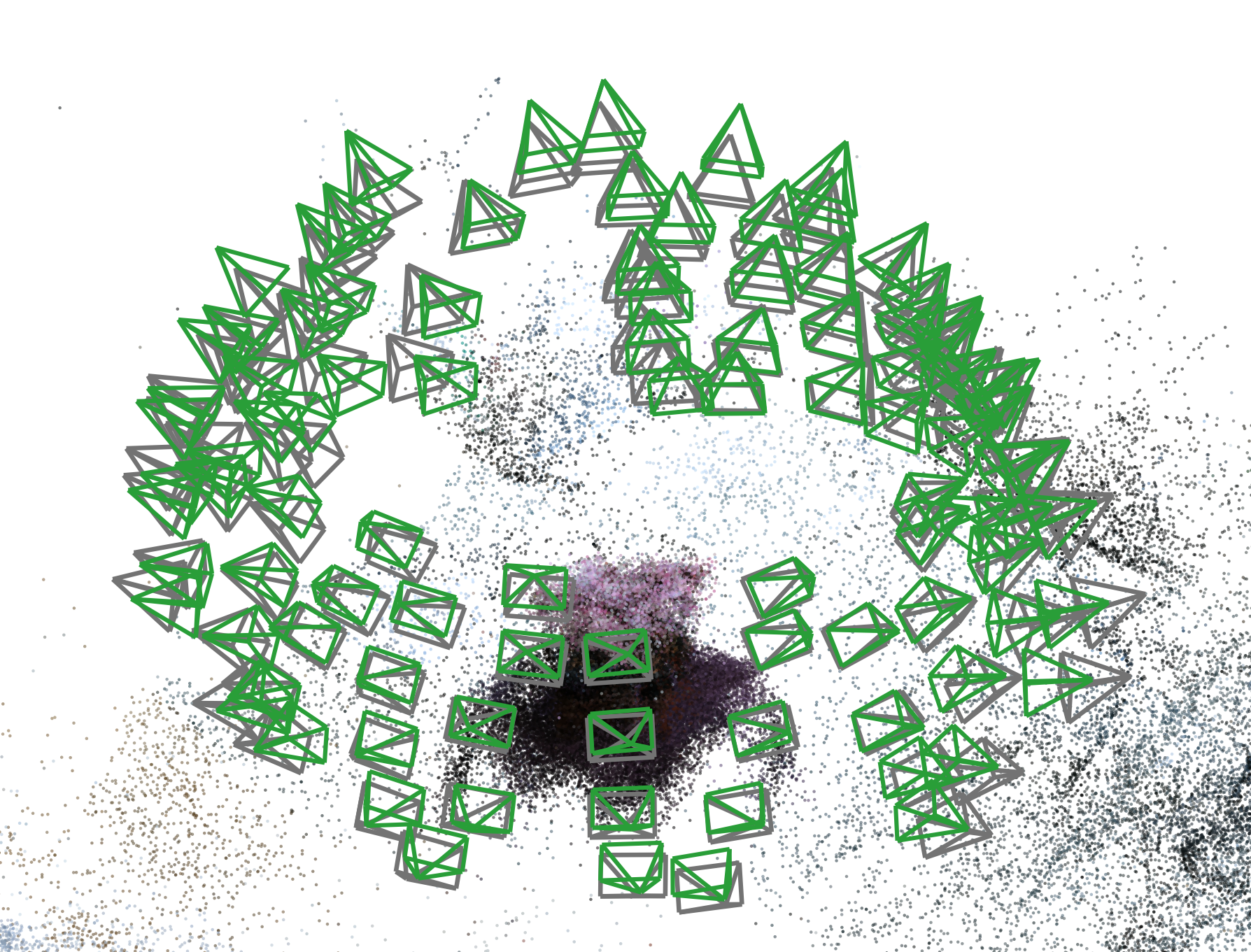}
        \caption{{\texttt{bonsai}: pitch bias}}
        \label{fig:failure_bonsai_3d}
    \end{subfigure}
    \caption{{\textbf{The \texttt{room} and \texttt{bonsai} failures on Mip-NeRF~360.} \ours{}'s reconstructions, with the estimated cameras in \cgreen{} and the ground-truth poses in \cgrey{} (every third camera in (b)). Only points with a reprojection error below $10$\,{px} are shown.}}
    \label{fig:failure_mip360}
\end{figure}

\paragraph{{The \texttt{room} failure.}}
\label{app:room}

{We attribute} the {\texttt{room}} failure to the view graph rather than to
scene size. Of \texttt{room}'s $311$ cameras and $5{,}829$ verified pairs, a spectral bipartition
separates $136$ cameras from the other $175$, and only $128$ pairs cross between the two. As a result, the two clusters are joined too weakly to fix
their relative pose, so each is reconstructed well internally while the two are placed
independently of one another {(\cref{fig:failure_room})}.

\paragraph{{The \texttt{bonsai} failure.}}
\label{app:bonsai}

\begin{figure}[!t]
    \centering
    \includegraphics[width=\linewidth]{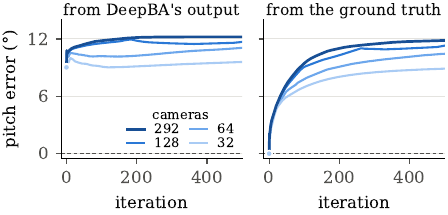}
    \caption{{\textbf{The \texttt{bonsai} pitch drift.} Median camera pitch error over \ours{}'s iterations on \texttt{bonsai} with $32$ to $292$ evenly spread cameras, started from \ours{}'s output (left) and from the ground truth (right).}}
    \label{fig:failure_bonsai}
\end{figure}

{Unlike \texttt{room}, \texttt{bonsai} has no view-graph bottleneck and almost no wrong matches. Instead, every camera is pitched up by about the same angle (${\sim}9${\textdegree{}}, \cref{fig:failure_bonsai_3d}), so cameras on opposite sides of the orbit are rotated against each other and most pairs miss the $10${\textdegree{}} threshold.

The tilt is a drift of \ours{}'s learned update. Even starting from the ground truth (zero error), \ours{} tilts the cameras to a median pitch error of $9.4${\textdegree{}} in $100$ iterations, and starting from its own output, it keeps tilting them further (\cref{fig:failure_bonsai}). Fewer cameras slow the drift but do not stop it. On $32$ evenly spread cameras, fewer than the $48$ {cameras of the largest training view graphs}, \ours{} started from the ground truth still reaches $7.0${\textdegree{}}. This example shows {a} limitation of DeepBA{.}}

\section{{Camera Localization}}
\label{app:localization}

\begin{table*}[t]
\centering
\renewcommand{\arraystretch}{0.95}
\caption{\textbf{{Camera} localization on 7-Scenes}~\cite{shotton2013scene}. {Median translation/rotation error (cm/\textdegree). The first row is HLoc~\cite{sarlin2019coarse} as published. The other rows share one frozen \ours{} map built from $200$ TrainSplit images. \textit{\textbf{Bold}/\underline{underline}: best/second best among them.}}}
\label{tab:localization_7scenes}
\begin{adjustbox}{max width=\linewidth}
\begin{tabular}{l|c|ccccccc}
\toprule
Method & \shortstack{{Intrinsics}\\required?} & Chess & Fire & Heads & Office & Pumpkin & Kitchen & Stairs \\
\midrule
HLoc~\cite{sarlin2019coarse}, ${\sim}4{,}000$-image map & \cmark & 2/0.79 & 2/0.87 & 2/0.92 & 3/0.91 & 5/1.12 & 4/1.25 & 6/1.62 \\
\midrule
\multicolumn{9}{l}{\textit{Same map, same queries -- only the registration step differs}} \\
HLoc~\cite{sarlin2019coarse}, our map & \cmark & {11/\underline{1.88}} & {10/\underline{1.89}} & {6/\underline{1.90}} & {\underline{9}/\underline{4.75}} & {16/3.85} & {10/4.37} & {\underline{37}/17.86} \\
PnP$+$RANSAC$+$GT $\mathbf{K}$ & \cmark & {12/1.91} & {11/\textbf{1.67}} & {6/2.04} & {10/\textbf{4.68}} & {15/\textbf{3.71}} & {10/\underline{4.21}} & {\underline{37}/\underline{17.39}} \\
PnPf$+$RANSAC & \xmark & {\underline{6}/\textbf{1.66}} & {\underline{5}/\textbf{1.67}} & {\underline{3}/2.01} & {11/4.84} & {\underline{10}/\underline{3.75}} & {\underline{9}/\textbf{3.96}} & {50/\textbf{17.35}} \\

\textbf{\ours{}} & \xmark & {\textbf{3}/\textbf{1.66}} & {\textbf{3}/\textbf{1.67}} & {\textbf{2}/\textbf{1.66}} & {\textbf{8}/4.80} & {\textbf{8}/3.86} & {\textbf{6}/4.30} & {\textbf{30}/18.30} \\
\midrule
\bottomrule
\end{tabular}%
\end{adjustbox}
\end{table*}

\begin{figure}[!ht]
  \centering
  \includegraphics[width=\linewidth]{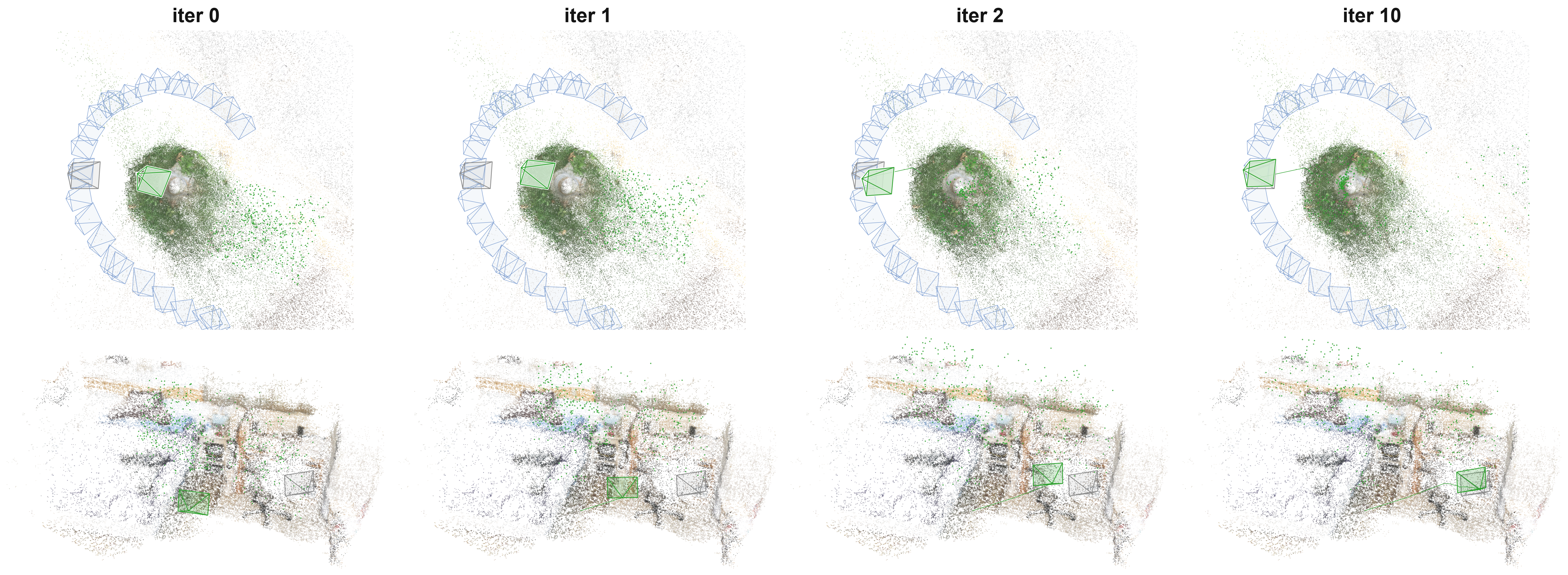}
  \caption{\textbf{Localization {reruns} the same refinement {iteration} in place.}
  With the reference map frozen {(its 3D points in their image colors, reference cameras in \cblue{})}, two example query views (rows) are registered
  across refinement iterations $0,1,2,10$ (columns){. From the mean pose of the reference cameras,} the query camera and its lifted 3D points (\cgreen{}) converge onto
  {its ground-truth camera (\cgrey{})} within ${\sim}10$ iterations.}
  \label{fig:localization_single}
\end{figure}

\begin{figure}[!ht]
  \centering
  \begin{subfigure}[b]{0.478\linewidth}
    \centering
    \includegraphics[height=3.3cm]{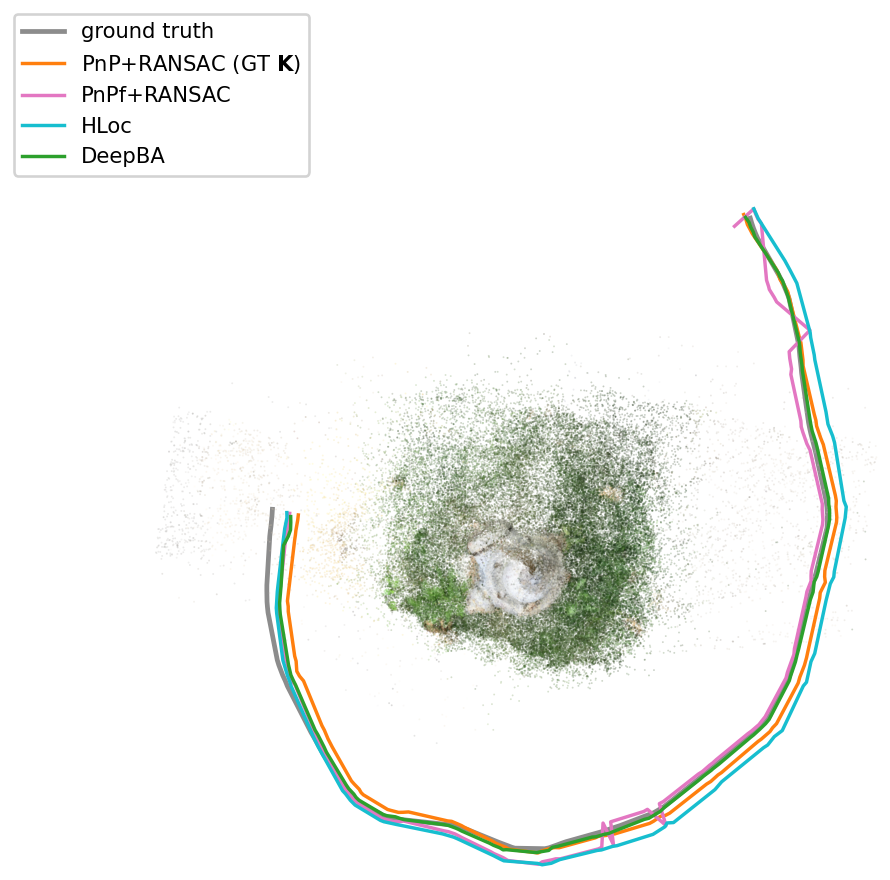}
    \caption{Object-centric orbit ({CO3Dv2}).}
    \label{fig:loc_comp_1}
  \end{subfigure}%
  \hfill%
  \begin{subfigure}[b]{0.522\linewidth}
    \centering
    \includegraphics[height=3.3cm]{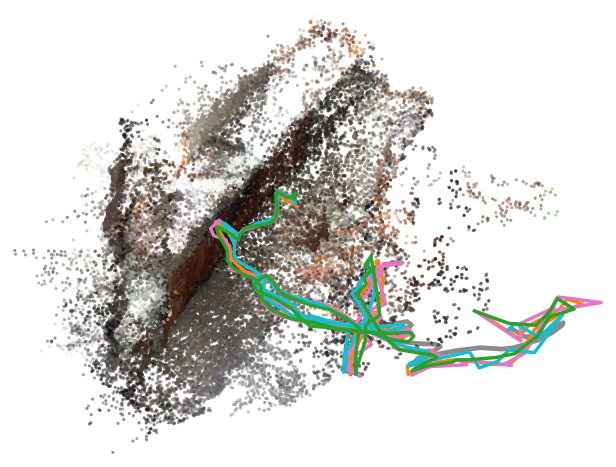}
    \caption{Indoor scene (ScanNet).}
    \label{fig:loc_comp_2}
  \end{subfigure}
  \caption{\textbf{Localization on a frozen \ours{} map.} Query-camera
  trajectories for four methods registering the same held-out queries against
  the same frozen \ours{} map, vs.\ ground truth
  (\textcolor{locgt}{\textbf{grey}}): \ours{}
  (\textcolor{locours}{\textbf{green}}), PnP+RANSAC given ground-truth
  intrinsics (\textcolor{locpnp}{\textbf{orange}}), focal-solving PnPf+RANSAC
  without known intrinsics (\textcolor{locpnpf}{\textbf{pink}}), and
  HLoc~\cite{sarlin2019coarse} (\textcolor{lochloc}{\textbf{cyan}}).}
  \label{fig:localization_compare}
\end{figure}

\ours{} also performs camera localization with no retraining and no architectural
change. We build a map from a set of reference views, freeze its cameras and
per-view depths, and register each query by rerunning the \emph{same} refinement
{iteration} with the reference updates zeroed, so no PnP is
involved (\cref{fig:localization_single}). This is possible because \ours{}'s
intermediate state is explicit geometry, absolute poses, per-view depths and
intrinsics, so any subset of it can simply be held fixed, whereas a feed-forward
regressor carrying only latent features would need an external solver.
\cref{fig:localization_compare} compares query trajectories against
PnP$+$RANSAC, PnPf$+$RANSAC and HLoc on the same frozen map, and
\cref{tab:localization_7scenes} reports 7-Scenes under the protocol of
HLoc~\cite{sarlin2019coarse}, mapping from the official TrainSplit, querying the
TestSplit, and reporting the median translation and rotation error per scene.
{Except for the published HLoc row, every} method in that comparison receives the same $200$-{image} \ours{} map rather
than the full ${\sim}4{,}000$ TrainSplit images, and both PnP$+$RANSAC and HLoc
additionally receive ground-truth query intrinsics. Localization is a by-product
of the reconstruction model rather than a main contribution of this paper.

\end{document}